%% file: main.tex
\documentclass{article}

\PassOptionsToPackage{numbers,compress}{natbib}
\usepackage[main,final]{neurips_2026}
\usepackage[most]{tcolorbox}
\usepackage[utf8]{inputenc}
\usepackage[T1]{fontenc}
\usepackage{hyperref}
\hypersetup{hidelinks}
\usepackage{url}
\usepackage{booktabs}
\usepackage{amsmath,amssymb,amsfonts}
\usepackage{nicefrac}
\usepackage{microtype}
\usepackage[table]{xcolor}
\usepackage{float}
\usepackage{needspace}
\usepackage{graphicx}
\usepackage{subcaption}
\usepackage{enumitem}
\usepackage{multirow}
\usepackage{placeins}
\usepackage{soul}
\usepackage{svg}
\usepackage{tabularx}
\usepackage{array}
\usepackage{makecell}
\usepackage{wrapfig}

\graphicspath{{fig/}}
\DeclareGraphicsExtensions{.pdf,.png,.jpg}

\definecolor{eviA}{RGB}{255, 249, 196}
\definecolor{eviB}{RGB}{204, 232, 255}
\definecolor{eviC}{RGB}{212, 245, 212}
\definecolor{eviD}{RGB}{255, 220, 235}

\usepackage{xcolor}
\definecolor{yifancolor}{HTML}{0072B2}

\newif\ifshowchanges
\showchangesfalse

\DeclareRobustCommand{\yifan}[1]{%
  \ifshowchanges
    {\color{yifancolor}#1}%
  \else
    #1%
  \fi
}
\title{Ask, Answer, and Detect: Role-Playing LLMs for Personality Detection with Question-Conditioned Mixture-of-Experts}

\author{%
  Yifan Lyu \\
  The Hong Kong University of \\
  Science and Technology (Guangzhou) \\
  \texttt{ylyu634@connect.hkust-gz.edu.cn}
  \And
  Liang Zhang\thanks{Corresponding author.} \\
  The Hong Kong University of \\
  Science and Technology (Guangzhou) \\
  \texttt{liangzhang@hkust-gz.edu.cn}
}

\begin{document}

\maketitle

\vspace{-0.4cm}
\begin{abstract}
Understanding human personality is crucial for web applications such as personalized recommendation and mental health assessment. Existing studies on personality detection predominantly adopt a ``posts $\rightarrow$ user vector $\rightarrow$ labels'' modeling paradigm, which encodes social media posts into user representations for predicting personality labels (\textit{e.g.}, MBTI labels). While recent advances in large language models (LLMs) have improved text encoding capacities, these approaches remain constrained by limited supervision signals due to label scarcity, and under-specified semantic mappings between user language and abstract psychological constructs. We address these challenges by proposing ROME, a novel framework that explicitly injects psychological knowledge into personality detection. Inspired by standardized self-assessment tests, ROME leverages LLMs’ role-play capability to simulate user responses to validated psychometric questionnaires. These generated question-level answers transform free-form user posts into interpretable, questionnaire-grounded evidence linking linguistic cues to personality labels, thereby providing rich intermediate supervision to mitigate label scarcity while offering a semantic reasoning chain that guides and simplifies the text-to-personality mapping learning. A question-conditioned Mixture-of-Experts module then jointly routes over post and question representations, learning to answer questionnaire items under explicit supervision. The predicted answers are summarized into an interpretable answer vector and fused with the user representation for final prediction within a multi-task learning framework, where question answering serves as a powerful auxiliary task for personality detection. 
Experiments on two real-world datasets show that ROME outperforms state-of-the-art baselines, achieving relative improvements of \textbf{13.60\%} and \textbf{19.72\%} on Kaggle and Pandora, respectively. Our code is available at https://github.com/JasmineLyf/ROME.
\end{abstract}

\vspace{-0.4cm}
\section{Introduction}
Understanding human personality is essential for explaining individual differences in cognition, emotion, and behavior and serves as the foundation for web applications such as personalized recommendation, mental-health assessment, and career development \cite{kernberg2016}. Among typological instruments, the Myers--Briggs Type Indicator (MBTI) remains one of the most widely adopted, categorizing individuals into sixteen personality types along four binary dimensions \cite{myers1998}. With the rapid growth of social media, individuals increasingly share their thoughts and experiences online, creating a rich source of behavioral signals for understanding human personality \cite{Pew2024SocialMediaUse}. There is now substantial evidence that digital traces and linguistic expressions on social media can reveal users' personality traits \cite{Schwartz2013PLOS,Youyou2015PNAS}. These developments have spurred the emerging task of \emph{personality detection}, which aims to infer personality labels (\textit{i.e.}, MBTI labels) from social media posts automatically and has attracted substantial research attention in recent years \cite{vstajner2020survey}.

% \begin{figure}[t]
\begin{wrapfigure}{r}{0.5\textwidth}
  \centering

  \begin{subfigure}[t]{\linewidth}
    \centering
    \includegraphics[width=0.98\linewidth]{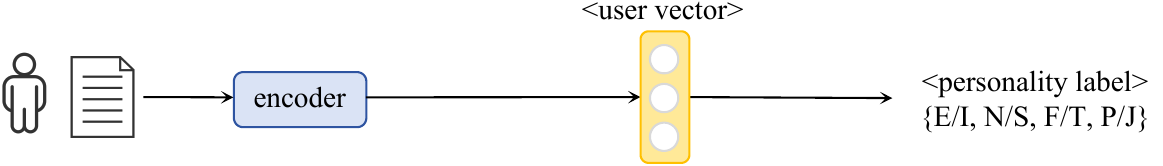}
    \caption{Traditional Deep Learning Models.}
    \label{fig:overview-a}
  \end{subfigure}

  \vspace{1pt}

  \begin{subfigure}[t]{\linewidth}
    \centering
    \includegraphics[width=0.98\linewidth]{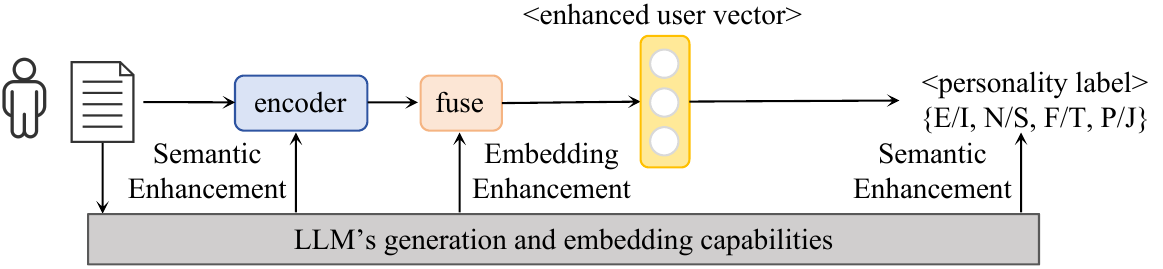}
    \caption{LLM-Enhanced Text Mapping Models.}
    \label{fig:overview-b}
  \end{subfigure}

  \vspace{1pt}

  \begin{subfigure}[t]{\linewidth}
    \centering
    \includegraphics[width=0.98\linewidth]{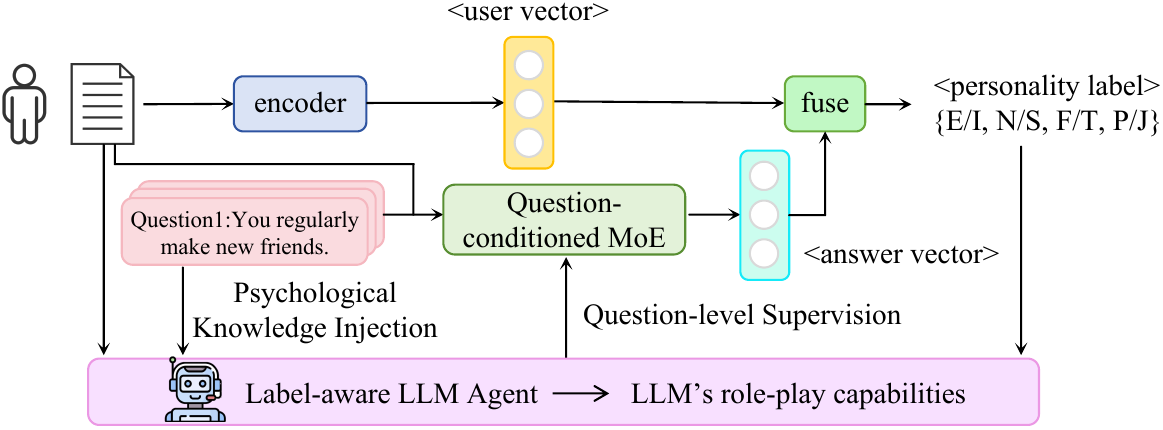}
    \caption{Our LLM Role-Play Enhanced, Questionnaire-Grounded Model.}
    \label{fig:overview-c}
  \end{subfigure}

  \caption{Overview of modeling paradigms for personality detection.}
  \label{fig:overview}
  \vspace{-0.6cm}
\end{wrapfigure}

These aforementioned studies typically formulate the task as a text classification problem following the ``posts $\rightarrow$ user vector $\rightarrow$ labels'' paradigm as illustrated in Figure~\ref{fig:overview}(a), where the crucial challenge is to learn a mapping function that encodes user posts into a single user representation. In this line of work, various strategies have been proposed. Traditional approaches primarily rely on handcrafted features, such as those derived from the Linguistic Inquiry and Word Count (LIWC) lexicon \cite{Pennebaker2015LIWC, Mairesse2007}. With the advent of deep learning, subsequent approaches have shifted from handcrafted features to neural architectures, which extract user representations from raw text automatically. For example, building on pretrained language models, some studies \cite{keh2019myers, jiang2020automatic} adopt BERT \cite{kenton2019bert} to encode user content, with strategies ranging from post-level encoding to concatenating multiple posts into one sequence. Some alternative methods \cite{yang2021mdt, zhu2022contrastive, yang2023ddgcn} adopt a cross-post perspective, using BERT to extract semantic features and graph neural networks to further enhance post representations. 
More recent extensions further enrich user representations via psychological guidance \cite{yang2021pqnet} or multi-view modeling \cite{zhu2024mvp}. For example, MvP \cite{zhu2024mvp} leverages multiple encoders with consistency regularization to help capture diverse perspectives of user posts.
Although these methods have achieved notable progress, their limited representational capacity often produces suboptimal embeddings that fail to fully exploit the rich personality signals embedded in user posts, especially when the posts are heterogeneous and noisy.
% Other extensions introduce additional psychological guidance or multi-view modeling to enrich user representations, for example by leveraging psychological questionnaires to capture personality-relevant cues from posts \cite{yang2021pqnet}, or by modeling users from multiple perspectives with a mixture-of-experts architecture and consistency regularization \cite{zhu2024mvp}. 

Recently, large language models (LLMs) have shown remarkable text understanding abilities and strong performance across a wide range of natural language processing tasks \cite{kaplan2020scaling,brown2020language,wei2022flan,wei2022cot} compared with traditional smaller models, which in turn has motivated their application to the personality detection task. The typical modeling paradigm is illustrated in Figure~\ref{fig:overview}(b). Specifically, TAE \cite{hu2024tae} leverages the generative capabilities of LLMs to produce multi-perspective textual analyses of posts and labels within a contrastive learning framework, thereby enriching post semantics and strengthening user representations. Building on TAE, ETM \cite{bi2025etm} further employs LLMs to exploit their strong embedding capacity, augmenting the user vector with long-text embeddings directly and strengthening the mapping between user vectors and personality labels. Extending this line,  LL4G \cite{shen2025ll4g} utilizes LLM embeddings to model inter-post relationships for dynamic graph construction, enhancing graph-based user representation learning. More recently, EmoPerso \cite{shen2025emoperso} incorporates LLM-based generative mechanisms including style-conditioned paraphrasing and contextual feature completion to construct and enrich emotion-aware user representations and achieves state-of-the-art performance. However, despite these advances, existing LLM-based approaches largely adhere to the traditional ``posts $\rightarrow$ user vector $\rightarrow$ labels'' paradigm, incorporating only limited augmentations. The semantic mapping from user language to abstract psychological constructs is achieved through simple text encoding and relies primarily on supervision from coarse-grained personality labels which are often scarce, costly to collect and prone to privacy concerns.

% Another line of work instead directly relies on inference-time LLM reasoning, including multi-turn questionnaire prompting, retrieval-augmented few-shot inference, and online hierarchical LLM-driven personality inference, as exemplified by PsyCoT, EERPD, and HIPPD \cite{yang2023psycot,li2025eerpd,chen2025hippd}. Since these methods require online LLM interaction or substantially heavier inference-time generation, they are fundamentally different from our setting, where LLM usage is confined to an offline stage and final prediction is performed by a compact supervised model.

To address this limitation, we observe that humans often infer their own personality by completing standardized self-assessment tests. Inspired by this process, we posit that structured, well-validated psychometric questionnaires provide a natural bridge between user language and abstract psychological constructs. Beyond coarse-grained personality labels, this questionnaires decompose each label into multiple semantically meaningful questions (\textit{i.e.}, psychometric items), and their fine-grained question–answer design provides rich intermediate supervision to mitigate label scarcity while offering a semantic reasoning chain that simplifies the text-to-personality mapping learning. 

Building on this insight, we propose \textbf{ROME} (\underline{RO}le-play enhanced question-conditioned \underline{M}ixture-of-\underline{E}xperts), an LLM role-play enhanced, questionnaire-grounded framework for personality detection, with its modeling paradigm illustrated in Figure~\ref{fig:overview}(c). Our key idea is to treat the standardized personality questionnaire as a structured retrieval source and design a label-aware LLM agent to role-play each user, generating question-level answers from their free-form posts. These answers constitute \emph{questionnaire-grounded evidence} that explicitly links linguistic cues to specific personality labels, bringing psychological knowledge directly into the modeling pipeline. We then design a question-conditioned Mixture-of-Experts (MoE) module that jointly routes over posts and question representations and learns to answer the psychometric questionnaire items under explicit supervision. These predicted answers are then summarized into an answer vector, a compact representation that captures the model’s question-by-question reasoning chain for personality detection. This vector is finally fused with the encoder’s user representation for prediction within a multi-task learning framework. In summary, our main contributions are as follows:

\begin{itemize}[leftmargin=1.5em, labelsep=0.5em]
    \item Conceptually, we are the first to leverage the strong role-play capabilities of LLMs to explicitly inject psychological knowledge into personality detection and formulate an interpretable, questionnaire-grounded modeling paradigm.
    \item Methodologically, we design a multi-task learning framework that treats question answering as a powerful auxiliary task for personality detection and develop a question-conditioned Mixture-of-Experts (MoE) architecture that learns to answer questionnaire items through a post–question joint routing mechanism.
    \item Experimentally, extensive results on two real-world datasets demonstrate that ROME significantly and consistently outperforms state-of-the-art baselines, achieving substantial improvements of $13.60\%$ on the \textsc{Kaggle} dataset and $19.72\%$ on the \textsc{Pandora} dataset.
\end{itemize}

\section{Method}
\label{sec:method}
\subsection{Problem Definition}
\label{sec:problem}
Personality detection from social media can be formulated as a \emph{user-level, multi-label} classification problem. Given a collection of posts authored by a user $u$, denoted as $\mathcal{P}_u = \{p_{u,1}, \dots, p_{u,N_u}\}$, where $N_u$ represents the total number of posts, the goal of this task is to infer the user’s \textit{m}-dimensional personality profile $\hat{\mathbf y}_u = \big(\hat y_u^{(1)}, \hat y_u^{(2)}, \cdots, \hat y_u^{(m)} \big)$ based on posts $\mathcal{P}_u$. 

Under the MBTI taxonomy, the personality dimension set is $\mathcal{M} = \{\mathrm{I/E}, \mathrm{S/N}, \mathrm{T/F}, \mathrm{P/J}\}$. Each dimension $m \in \mathcal{M}$ consists of two opposing trait poles, and the ground-truth label $y_u^{(m)} \in \{0, 1\}$ indicates which pole the user $u$ aligns with. The objective is therefore to learn the mapping function from $\mathcal{P}_u$ to $\hat{\mathbf y}_u$: 
\[
f:\ \mathcal{P}_u \ \longmapsto\ \hat{\mathbf y}_u = \big(\hat y_u^{(m)}\big)_{m\in\mathcal{M}}.
\]

% where each $y_u^{(m)}$ denotes a one-hot encoded label corresponding to dimension $m \in \mathcal{M}$. 

\subsection{Model Overview}
\label{subsec:overview}
\begin{figure}[h]
    \centering
    \includegraphics[width=1.0\textwidth]{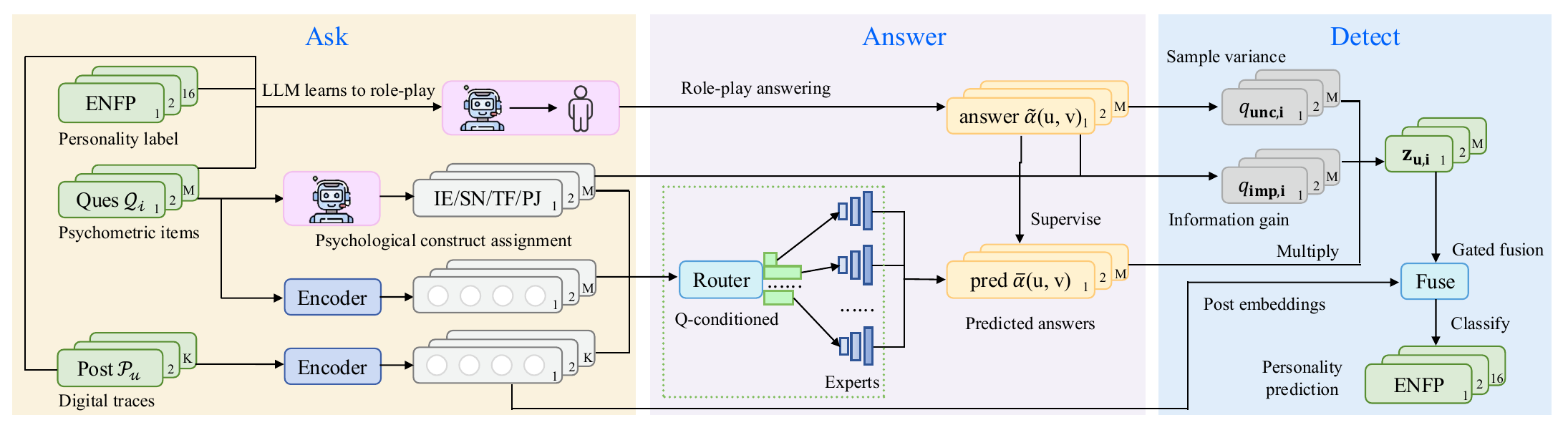} 
    \caption{The overall pipeline of ROME.}
    \label{fig:pipeline}
\vspace{-0.5cm}
\end{figure}

To facilitate and ease the learning process from the free-form user language space $\mathcal{P}_u$ to the abstract psychological construct space $\hat{\mathbf y}_u$, we incorporate well-validated psychometric questionnaires as a natural intermediate bridging layer, drawing inspiration from the standardized self-assessment procedure widely adopted in human personality measurement in psychology studies. Building on this insight, we propose \textbf{ROME} (\underline{RO}le-play enhanced question-conditioned \underline{M}ixture-of-\underline{E}xperts), an LLM role-play enhanced, questionnaire-grounded framework for personality detection, with its overall architecture illustrated in Figure \ref{fig:pipeline}. 

Conceptually, ROME comprises three key stages: \emph{Ask}, \emph{Answer}, and \emph{Detect}. Specifically, in the Ask stage, a role-play LLM agent is instructed to ``take on the perspective of the user" and produce answers to each questionnaire item \footnote{Henceforth, we use the terms ``questionnaire item(s)'' and ``question(s)'' interchangeably for convenience and readability throughout this paper.} based on \yifan{the training user's posts and the corresponding
training label}. %the user’s posts and the corresponding personality label.
These questionnaire-grounded evidence serves as a structured and interpretable bridge between raw linguistic expressions and psychological constructs. Importantly, this evidence generation process is performed offline during training only and is no longer required at inference time. Then, in the Answer stage, we enable the model to reproduce these evidence directly from the user posts and the questionnaire item texts, thereby establishing a psychology-informed semantic reasoning chain that simplifies the text-to-personality mapping learning. To account for the heterogeneity across psychometric questionnaire items, we introduce a question-conditioned Mixture-of-Experts (MoE) module, where each expert specializes in a particular type of psycholinguistic cue and the gating network selects the most relevant experts for answering each question. Finally, in the Detect stage, the questionnaire-grounded evidence is aggregated with the user representation learned from posts to perform the final prediction, preserving the expressive capacity of the post encoder while injecting a concise and interpretable decision pathway grounded in standardized psychological knowledge.

\subsection{Ask: Role-Play-Based Questionnaire Response Generation}
\label{subsec:ask}

% \begin{figure}[h]
\begin{wrapfigure}{r}{0.5\textwidth}
    
    \centering\includegraphics[width=0.98\linewidth]{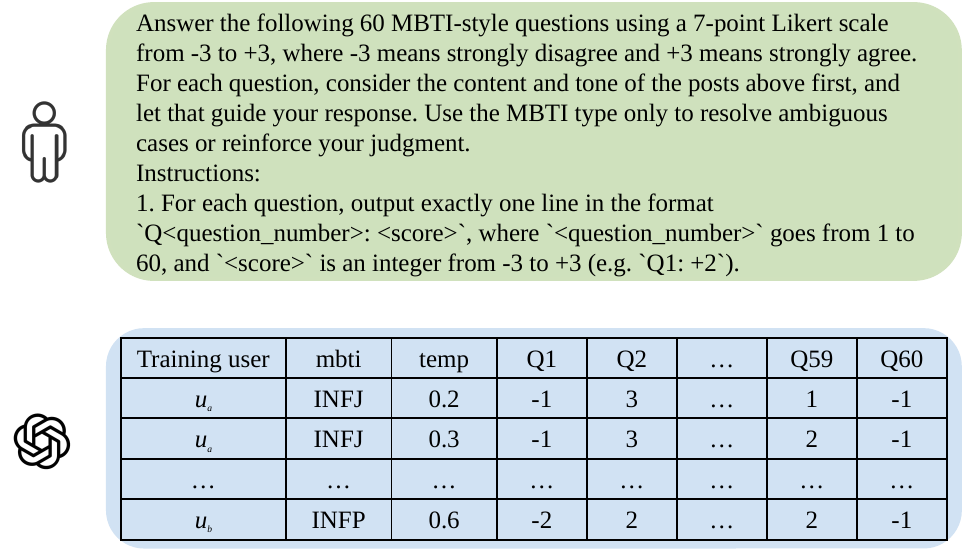} 
    \caption{Prompt design for role-play-based questionnaire response generation.}
    \label{fig:prompt}
    \vspace{-0.5cm}
% \end{figure}
\end{wrapfigure}

To ease the learning of the mapping function $f(\cdot)$, we introduce standardized psychometric questionnaires to decompose each personality label into multiple semantically meaningful questions, thereby providing rich, fine-grained intermediate supervision signals for the model. Ideally, one could present these self-assessment tests directly to each user $u$ and collect their responses to obtain such supervision; however, this procedure is costly, intrusive, and often impractical. Motivated by the recent success of LLMs in simulating human behaviors across diverse contexts \cite{abbasiantaeb2024let, xie2024can}, we propose an LLM agent to role-play each user with given characteristics (\textit{i.e.}, user posts $\mathcal{P}_u$ and personality label $\mathbf{y}_u$) and produce question-level answers. The posts provide evidence of the user’s linguistic patterns and behavioral tendencies, while the label supplies a stable trait orientation; conditioning on both factors enables the LLM agent to more accurately approximate how the user would respond in a self-assessment setting. Here, it is important to note that this generation process is performed offline during training only to augment the training dataset and is not required at inference time.

Formally, for each user $u$ with posts $\mathcal{P}_u = \{p_{u,1}, \dots, p_{u,N_u}\}$ and each questionnaire item $Q_i \in \mathcal{Q}$, we apply the prompt shown in Figure~\ref{fig:prompt} to simulate the user’s response. The LLM agent outputs a score on an ordered scale that approximates how user $u$ would answer $Q_i$. Since LLM generation is inherently stochastic, we draw $T$ samples and average them to reduce sampling variance and obtain a more robust estimate, as follows.
\begin{equation}
\tilde{a}_{u,i}\;=\;\frac{1}{T}\sum_{t=1}^{T} a^{(t)}_{u,i}, \quad u\in\mathcal{U},\ Q_i \in \mathcal{Q},
\label{eq:ask-avg}
\end{equation}
where $a^{(t)}_{u,i}$ denotes the $t$-th sampled answer for the pair $(u, Q_i)$. These answers bring psychological knowledge explicitly into the modeling framework.

\subsection{Answer: Question-Conditioned Mixture-of-Experts Framework}
\label{subsec:answer}
% first main task -> why moe structure -> the rounting design

The goal of the Answer stage is to enable the model to predict the question-level answers produced in the Ask stage directly from the user’s posts which are accessible during inference. By learning to reproduce these answers, the model internalizes the psychology-informed semantic reasoning chain, thereby facilitating the text-to-personality mapping process.

A straightforward approach is to train a neural encoder to map a user’s posts to each question-level answer. However, different questionnaire items probe distinct psychological constructs (\textit{e.g.}, emotional stability vs. social expressiveness), and the linguistic cues associated with these constructs may also differ substantially across user posts. Therefore, a single shared encoder tends to conflate heterogeneous semantic signals, which leads to entangled representations and suboptimal generalization across questions. To address these limitations, we propose a question-conditioned Mixture-of-Experts (MoE) architecture. Instead of relying on a single shared encoder, the MoE structure allows different experts to specialize in capturing distinct psychological constructs and their associated linguistic cues, while still enabling shared experts to model knowledge that is broadly useful across questions.
% supporting both specialization and knowledge sharing naturally. 

Formally, we assume a set of $K$ experts $\mathcal{E} = \{E_{1}, \dots, E_{K}\}$, each parameterized as a neural encoder. Given a user $u$ and a specific question $Q_i$, the model should first identify the experts that are most informative to generate the corresponding answer. This selection is governed by a routing mechanism. Conceptually, effective routing requires understanding \textit{who} is being assessed (the user’s linguistic profile), \textit{what} is being asked (the content of the question), and \textit{which} psychological construct the question is designed to measure. Guided by this intuition, we introduce a lightweight question-conditioned router to dynamically select and aggregate experts as follows.
\begin{equation}
\begin{aligned}
\mathbf{x}_{u,i} = \left[ \mathbf{v}_u;\ \mathbf{q}_i;\ \mathbf{e}_{m_i} \right], \,\,
\mathbf{g}_{u,i} = \text{softmax}\left(\text{MLP}(\mathbf{x}_{u,i}) \right).
\end{aligned}
\label{eq:joint-x-answer}
\end{equation}
where $\mathbf{v}_u$ and $\mathbf{q}_i$ denote the text embeddings of the user posts $\mathcal{P}_u$ and the question $Q_i$ respectively, computed by the shared textual encoder $t(\cdot)$. The vector $\mathbf{e}_{m_{i}}$ is a one-hot indicator specifying the psychological construct that question $Q_i$ is designed to measure (\textit{e.g.}, I/E or S/N personality dimension). In our implementation, we employ an LLM to obtain the construct assignment once, and keep this assignment fixed throughout training and evaluation. The prompt design used by the LLM agent is provided in Appendix~\ref{app:imple}.

Each expert $E_k$ is a small MLP-based expert function denoted as $f_k(\cdot)$. The final answer is then predicted as:
\begin{equation}
\hat{a}_{u,i} = \sum_{k=1}^{K} g^{(k)}_{u,i}\, \cdot f_k(\mathbf{x}_{u,i}).
\label{eq:moe-mix-answer}
\end{equation}
This component is trained to predict the answer for a given user and psychometric question pair $(u, Q_i)$ using the role-played target $\tilde a_{u,i}$ as supervision, and optimized via a regression objective function as follows.
\begin{equation}
\mathcal{L}_{q} = \frac{1}{|\mathcal{U}| \cdot |\mathcal{Q}|} \sum_{u \in \mathcal{U}}\sum_{Q_i \in \mathcal{Q}} \ell \left(\hat a_{u,i},\,\tilde a_{u,i} \right),
\label{eq:lq-train}
\end{equation}
where $\ell(\cdot,\cdot)$ denotes a robust L1 regression loss. During inference, the predicted $\hat{a}_{u,i}$ serves as a strong psychologically grounded evidence for final personality detection. 

% 改到这里了
\subsection{Detect: Psychological Evidence-Guided Personality Prediction}
\label{subsec:detect}

\subsubsection{Psychological Evidence Modulation}
% key logic?
Given the predicted $\hat{a}_{u,i}$ as psychologically grounded evidence, one may simply concatenate them to form an evidence vector for personality prediction. However, two critical challenges remain. First, the ``ground truth'' answers used to supervise $\hat{a}_{u,i}$ are obtained through LLM role-play generation. Due to the stochastic nature of LLM outputs, the augmented answers may contain noise, leading to varying levels of uncertainty across questions. Second, not all questions in a questionnaire are equally informative for personality inference. Some questions provide strong diagnostic signals compared to others. Therefore, we propose to adaptively weight the contribution of each predicted answer based on both its reliability and its importance when forming the final answer-based representation. 

\paragraph{Reliability Score.}
Specifically, when estimating the reliability, our intuition is that if the responses of a question exhibit high variance across the $T$ sampled role-play generations, the corresponding evidence is more likely to be unreliable. Formally, we can define it as follows.
\begin{equation}
\begin{aligned}
q_{\mathrm{unc},i} = \frac{1}{|\mathcal{U}|}\sum_{u\in\mathcal{U}} \text{Var}\left(\{a^{(t)}_{u,i}\}_{t=1}^{T}\right), \,\, 
q_{\mathrm{rel},i} = 1 - \text{Norm}\left(q_{\mathrm{unc},i}\right).
\end{aligned}
\label{eq:qunc-def}
\end{equation}
where $q_{\mathrm{rel},i}$ denotes the reliability score associated with question $Q_i$,  
and $\text{Norm}(\cdot)$ represents the min–max normalization operator. 

\paragraph{Importance Score.}
While for the importance, our intuition is that each question is primarily designed to probe a specific psychological construct (\textit{e.g.}, I/E or S/N personality dimension). Thus, if the LLM-generated answers for a question exhibit stronger separability and induce larger between-class shifts with respect to its intended construct, then that question should be considered more important for personality detection. Formally, for each question $Q_i$, we first retrieve its one-hot indicator vector $\textbf{e}_{m_i}$ which is a fixed mapping introduced in the previous section, and denote the corresponding psychological construct by $m \in \mathcal{M}$ (\textit{e.g.}, $m=\text{I/E}$). The importance is then defined as:
\begin{equation}
\begin{aligned}
\mu_{i, m}^{+} = \frac{1}{|\mathcal{U}_{m}^{+}|}\sum_{u\in\mathcal{U}_{m}^{+}}\tilde a_{u,i}, \,\,
\mu_{i, m}^{-} = \frac{1}{|\mathcal{U}_{m}^{-}|}\sum_{u\in\mathcal{U}_{m}^{-}}\tilde a_{u,i}, \,\,
q_{\mathrm{imp},i} = \left|\mu_{i, m}^{+} - \mu_{i, m}^{-}\right|,
\end{aligned}
\label{eq:class-means}
\end{equation}
where $\mathcal{U}_m^{+}=\{u\mid y_u^{(m)}=1\}$ and $\mathcal{U}_m^{-}=\{u\mid y_u^{(m)}=0\}$, $q_{\mathrm{imp},i}$ is the importance score associated with question $Q_i$, which is min–max normalized to the range $[0,1]$.

\paragraph{Adaptive Weight.}
For each question $Q_i$, the final weight $w_i$ is computed under the intuition that more important and more reliable questions should contribute more to the final prediction, as follows.
\begin{equation}
w_i = q_{\mathrm{imp},i} \cdot q_{\mathrm{rel}, i}.
\label{eq:triple-prior}
\end{equation}
\yifan{Both reliability and importance scores are computed using training users only, and the resulting question weights are kept fixed during validation and testing.}
And the final answer-based representation is obtained by:
\begin{equation}
\textbf{s}_u = \textbf{w} \odot \hat{\textbf{a}}_{u}.
\end{equation}
where $\hat{\textbf{a}}_{u}$ denotes the vector of predicted question-level answers and $\odot$ denotes element-wise multiplication.

\subsubsection{Personality Prediction} Each questionnaire item \(Q_i\) is primarily designed to probe a specific psychological construct and contributes only minimally to others. Directly applying question-level signals without regard to their associated constructs may introduce irrelevant evidence and increase noise. To prevent such cross-construct interference, we apply construct-specific binary masks to ensure that each construct only integrates evidence from its corresponding questions. Specifically, for each \(m \in \mathcal{M}\) (\textit{e.g.}, $m=\text{I/E}$), we define the mask vector \(\boldsymbol{\mu}^{(m)} \in \{0,1\}^{|\mathcal{Q}|}\) by:
\begin{equation}
\boldsymbol{\mu}_i^{(m)} =
\begin{cases}
1, & \arg\max(\textbf{e}_{m_i})=m, \\[2pt]
0, & \text{otherwise}.
\end{cases}
\label{eq:mask-def}
\end{equation}

And the construct-specific answer representation is then computed by applying the obtained binary mask that preserves only the evidence associated with $m$ as follows. 
\begin{equation}
\mathbf{s}^{(m)}_{u} = \boldsymbol{\mu}^{(m)} \odot \mathbf{s}_u.
\label{eq:masked-evidence}
\end{equation}

The post-derived representation $\textbf{v}_u$ captures broad linguistic patterns across a user’s posts and is highly expressive, but it is not explicitly aligned with psychological constructs. In contrast, the construct-specific evidence $\mathbf{s}^{(m)}_{u}$ is questionnaire-grounded, providing a semantic and interpretable reasoning pathway to psychological constructs. However, its quality can vary due to differences in diagnostic strength across questionnaire items and the uncertainty inherent in role-play supervision. Therefore, the two sources offer complementary advantages and with varying reliability. Their contributions should be adaptively balanced during personality prediction. Specifically, we introduce a lightweight gating network that learns to dynamically weight the contributions of $\textbf{v}_u$ and $\mathbf{s}^{(m)}_{u}$ as: 
\begin{equation}
\begin{aligned}
\boldsymbol{\gamma}^{(m)}_u = 
\sigma\,\Bigl(\mathrm{MLP}^{(m)}_{\mathrm{gate}}\bigl(\mathbf{v}_u \,\|\, \mathbf{s}^{(m)}_u\bigr)\Bigr), \,\,
\mathbf{z}^{(m)}_u = \boldsymbol{\gamma}^{(m)}_u \odot \mathbf{v}_u + \bigl(\mathbf{1}-\boldsymbol{\gamma}^{(m)}_u\bigr)\odot \mathbf{s}^{(m)}_u.
\end{aligned}
\label{eq:gate}
\end{equation}

Subsequently, the personality prediction for a given \(m \in \mathcal{M}\) is produced by a lightweight construct-specific classifier as follows.
\begin{equation}
\hat y^{(m)}_u = \sigma\left(\mathrm{MLP}^{(m)}_{\mathrm{cls}}\bigl(\mathbf{z}^{(m)}_u\bigr)\right),
\label{eq:head}
\end{equation}
which is optimized using a binary classification loss function:
\begin{equation}
\mathcal{L}_{\mathrm{cls}} = \frac{1}{|\mathcal{U}| \cdot |\mathcal{M}|} \sum_{u \in \mathcal{U}} \sum_{m\in\mathcal{M}} \mathrm{BCE}\left(\hat y_u^{(m)},\,y_u^{(m)}\right). 
\label{eq:lcls-train}
\end{equation}

Together with the answer prediction loss defined in Equation~\eqref{eq:lq-train}, the model is finally trained in a multi-task learning framework with the following joint objective function:
\begin{equation}
\mathcal{L} = \lambda_{q}\,\mathcal{L}_{q} + \lambda_{cls}\,\mathcal{L}_{cls},
\label{eq:joint-train}
\end{equation}
where $\lambda_{q}$ and $\lambda_{cls}$ are hyperparameters that balance the contributions of the answer prediction and personality classification tasks, respectively. 

\noindent\textbf{Remark.} During training, we first pre-train the MoE module with the answer prediction objective (\textit{i.e.}, Equation~\eqref{eq:lq-train}) to stabilize the question-aligned reasoning space, and then fine-tune the entire framework end-to-end using the joint objective in Equation~\eqref{eq:joint-train}. At inference time, the user posts are encoded once, and the question-conditioned MoE module produces psychologically grounded evidence $\{\hat a_{u,i}\}$, which is then fused with the encoded post representations to generate the final personality prediction.

\section{Experiment}

% \subsection{Experimental Setting}
% \paragraph{Experiment Overview.}
% We evaluate ROME from five perspectives: overall performance, component ablations, comparison with inference-time LLM methods, data efficiency, and backbone sensitivity. Additional analyses are provided in Appendix~\ref{app:param-analysis} and Appendix~\ref{app:case-study}.

\noindent\textbf{Experiment Overview.}
We first report the overall comparison results with existing methods in Section~\ref{subsec:overall-results}, followed by component ablations in Section~\ref{subsec:ablation}. We then compare ROME with representative inference-time LLM methods in Section~\ref{subsec:llm-inference} and examine its data efficiency in Section~\ref{subsec:data-efficiency}. Additional experiments on backbone LLMs and text encoders are provided in Appendix~\ref{app:backbone}. Parameter analyses, including questionnaire length and the number of experts, are presented in Appendix~\ref{app:param-analysis}. A three-part qualitative case study covering question–post evidence chains, learned question weights, and expert activation patterns is included in Appendix~\ref{app:case-study}.

\noindent\textbf{Dataset.} In line with prior work~\cite{yang2021mdt,yang2023ddgcn,hu2024tae}, we conduct our evaluation on two widely used personality datasets: \textsc{Kaggle}
% \footnote{https://www.kaggle.com/datasnaek/mbti-type} 
and \textsc{Pandora}~\cite{gjurkovic-etal-2021-pandora}.
% \footnote{https://psy.takelab.fer.hr/datasets/all}
% Both datasets adopt the MBTI taxonomy with four binary dimensions: Introversion vs.\ Extroversion (I/E), Sensing vs.\ iNtuition (S/N), Thinking vs.\ Feeling (T/F), and Perception vs.\ Judging (P/J). 
Given pronounced class imbalance, we report Macro-F1 per dimension and the overall score as the average of the four Macro-F1s. 
% Unless otherwise specified, all  Macro-F1 scores are reported on a 0--100 scale. 
Following~\cite{yang2023ddgcn}, we shuffle users and partition data into train/validation/test with a 60/20/20 split. Dataset statistics are summarized in Table~\ref{tab:dataset-stats} in Appendix~\ref{app:imple}. 

\noindent\textbf{Baselines.} We compare our approach against \textbf{16} strong baseline methods, including traditional and basic neural models (XGBoost~\cite{chen2016xgboost}, BiLSTM~\cite{tadesse2017personality}, BERTconcat~\cite{jiang2020automatic}, and BERTmean~\cite{keh2019myers}), deep and structure-enhanced models (AttRCNN~\cite{Xue2018AppliedIntelligence}, AttnSeq~\cite{lynn-etal-2020-hierarchical}, Transformer-MD~\cite{yang2021mdt}, TrigNet~\cite{yang-etal-2021-psycholinguistic}, PQ-Net~\cite{yang2021pqnet}, D-DGCN and D-DGCN+$\ell_{0}$~\cite{yang2023ddgcn}, and MvP~\cite{zhu2024mvp}), and recent LLM-assisted methods (TAE~\cite{hu2024tae}, ETM~\cite{bi2025etm}, LL4G~\cite{shen2025ll4g}, and EmoPerso~\cite{shen2025emoperso}). For recent LLM-assisted baselines that support different backbone LLMs (\textit{e.g.}, LL4G and EmoPerso), we evaluate their \texttt{GPT-4o}-based variants in our experiments to ensure a fair comparison. The implementation details are provided in Appendix~\ref{app:imple}. 

\begin{table*}[t]
\centering
\caption{Performance comparison on the \textsc{Kaggle} and \textsc{Pandora} datasets.}
\label{tab:main_results}
\footnotesize
\setlength{\tabcolsep}{2.2pt}
\renewcommand{\arraystretch}{0.92}
\resizebox{\textwidth}{!}{%
\begin{tabular}{lcccccccccc}
\toprule
\multirow{2}{*}{\textbf{Methods}}
& \multicolumn{5}{c}{\textbf{\textsc{Kaggle}}}
& \multicolumn{5}{c}{\textbf{\textsc{Pandora}}} \\
\cmidrule(lr){2-6}\cmidrule(lr){7-11}
& \textbf{I/E} & \textbf{S/N} & \textbf{T/F} & \textbf{P/J} & \textbf{Avg}
& \textbf{I/E} & \textbf{S/N} & \textbf{T/F} & \textbf{P/J} & \textbf{Avg} \\
\midrule
XGBoost        & 56.67 & 52.85 & 75.42 & 65.94 & 62.72 & 45.99 & 48.93 & 63.51 & 55.55 & 53.50 \\
BiLSTM         & 57.82 & 57.87 & 69.97 & 57.01 & 60.67 & 48.01 & 52.01 & 63.48 & 56.21 & 54.93 \\
BERTconcat     & 58.33 & 53.88 & 69.36 & 60.88 & 60.61 & 54.22 & 49.15 & 58.31 & 53.14 & 53.71 \\
BERTmean       & 64.05 & 57.82 & 77.06 & 65.25 & 66.04 & 56.60 & 48.71 & 64.70 & 56.07 & 56.52 \\
\midrule
AttRCNN        & 59.74 & 64.08 & 78.77 & 66.44 & 67.25 & 48.55 & 56.19 & 64.39 & 57.26 & 56.60 \\
AttnSeq        & 65.43 & 62.15 & 78.05 & 63.92 & 67.39 & 56.98 & 54.78 & 60.95 & 54.81 & 56.88 \\
Transformer-MD & 66.08 & 69.10 & 79.19 & 67.50 & 70.47 & 55.26 & 58.77 & 69.26 & 60.90 & 61.05 \\
TrigNet        & 69.54 & 67.17 & 79.06 & 67.69 & 70.86 & 56.69 & 55.57 & 66.38 & 57.27 & 58.98 \\
D-DGCN         & 68.41 & 65.66 & 79.56 & 67.22 & 70.21 & 61.55 & 55.46 & \underline{71.07} & 59.96 & 62.01 \\
D-DGCN+$\ell_{0}$ & 69.52 & 67.19 & 80.53 & 68.16 & 71.35 & 59.98 & 55.52 & 70.53 & 59.56 & 61.40 \\
PQ-Net         & 68.94 & 67.65 & 79.12 & 69.57 & 71.32 & 57.07 & 55.26 & 65.64 & 58.74 & 59.18 \\
MvP            & 67.68 & 69.89 & 80.99 & 68.32 & 71.72 & 60.08 & 56.99 & 69.12 & 61.19 & 61.85 \\
\midrule
TAE            & 70.90 & 66.21 & 81.17 & 70.20 & 72.12 & 62.57 & 61.01 & 69.28 & 59.34 & 63.05 \\
ETM            & 68.97 & 71.21 & \underline{86.19} & \textbf{84.78} & 77.79 & \underline{68.57} & 64.91 & 66.07 & 63.53 & 65.77 \\
LL4G           & 79.05 & \underline{78.27} & 86.03 & 76.91 & 80.06 & 66.24 & 67.55 & \underline{71.30} & 66.91 & 68.00 \\
EmoPerso       & \underline{81.12} & 78.23 & 84.45 & 79.21 & \underline{80.75} & 68.45 & \underline{67.98} & 70.10 & \underline{69.77} & \underline{69.08} \\
\midrule
\textbf{ROME} (BERT)
& $\mathbf{92.24{\pm}0.47}$
& $\mathbf{95.90{\pm}0.17}$
& $\mathbf{97.45{\pm}0.12}$
& $\underline{81.32{\pm}0.28}$
& $\mathbf{91.73{\pm}0.11}$
& $\mathbf{74.93{\pm}0.24}$
& $\mathbf{95.45{\pm}0.04}$
& $\mathbf{82.21{\pm}0.09}$
& $\mathbf{78.22{\pm}0.07}$
& $\mathbf{82.70{\pm}0.11}$ \\
\midrule
\textbf{Improvement} & 13.71\% & 22.52\% & 13.06\% & -4.08\% & 13.60\% & 9.28\% & 40.41\% & 15.30\% & 12.11\% & 19.72\% \\
\bottomrule
\end{tabular}
}
\vspace{-0.4cm}
\end{table*}

\subsection{Overall Results}
\label{subsec:overall-results}
The main results on the two datasets are provided in Table~\ref{tab:main_results}. We observe that traditional feature-based models and standard PLM baselines perform poorly on both datasets. For example, BERTmean achieves only $66.04$ and $56.52$ average Macro-F1 on Kaggle and Pandora, respectively. In contrast, recent LLM-enhanced baselines deliver substantially stronger performance. In particular, EmoPerso achieves the strongest baseline results, reaching $80.75$ and $69.08$ average Macro-F1 on the two datasets. ROME further improves over these strong LLM-augmented baselines, achieving $91.73$ and $82.70$  Macro-F1 on Kaggle and Pandora, corresponding to relative improvements of $13.60\%$ and $19.72\%$ over EmoPerso. 

We further observe that ROME consistently and significantly outperforms ETM even though it relies only on a small-scale text encoder (\texttt{BERT-base-uncased}), whereas ETM exploits long-text embeddings from a much larger \texttt{Meta-Llama-3-8B-Instruct} model (about $70\times$ larger than \texttt{BERT-base-uncased}). This advantage mainly stems from ROME's questionnaire-grounded architecture, thereby providing substantially richer intermediate supervision than coarse-grained personality labels alone. Additional experiments in Appendix \ref{app:backbone} demonstrate that substituting BERT with LLaMA will further increase the performance from $91.73$ to $93.61$ on the \textsc{Kaggle} dataset.

\subsection{Ablation Studies}
\label{subsec:ablation}
To assess the contribution of each component in our ROME framework, we conduct an ablation study on the \textsc{Kaggle} dataset, as shown in Table~\ref{tab:model-variants} and Table~\ref{tab:moe-ablation}. 

\noindent\textbf{Benefits of adaptive weighting.} We first evaluate the impact of the question-level weighting mechanism. Removing the reliability- and importance-based weights (ROME$_{\text{w/o q-weighting}}$) reduces the average Macro-F1 from $91.73$ to $88.47$. This degradation reflects the heterogeneity across questionnaire items: different items vary in their uncertainty in LLM-simulated answers and in their discriminative power for distinguishing personality classes. 

\noindent\textbf{Benefits of gated fusion.} We next examine the fusion mechanism between post-derived and questionnaire-grounded representations. Replacing the learnable gated fusion with a simple unweighted average (ROME$_{\text{w/o gated fusion}}$) yields a similar decline in performance. This result indicates that the relative contributions of posts and questionnaire evidence should be adaptively determined.
% rather than fixed, as their reliability can vary across users and across personality dimensions.

% \begin{table}[t]
\begin{wraptable}{r}{0.52\textwidth}
\centering
\caption{Results of the ablation study on the \textsc{Kaggle} dataset.}
\label{tab:model-variants}
\small
\setlength{\tabcolsep}{4pt}
\renewcommand{\arraystretch}{0.9}
\begin{tabular}{lccccc}
\toprule
\textbf{Methods} & \textbf{I/E} & \textbf{S/N} & \textbf{T/F} & \textbf{P/J} & \textbf{Avg} \\
\midrule
ROME$_{\text{w/o q-weighting}}$  & \underline{90.13} & \underline{94.82} & \underline{96.36} & 72.55 & 88.47 \\
ROME$_{\text{w/o gated fusion}}$ & 89.03 & 94.45 & \underline{96.36} & \underline{74.16} & \underline{88.50} \\
ROME$_{\text{w/o posts}}$        & 86.56 & 71.21 & 94.18 & 48.51 & 75.12 \\
ROME$_{\text{w/o evidence}}$     & 42.33 & 92.35 & 75.11 & 53.27 & 65.77 \\
\yifan{ROME$_{\text{label-only Ask}}$} &
\yifan{47.02} & \yifan{91.53} & \yifan{79.02} & \yifan{56.09} & \yifan{68.42} \\
\yifan{ROME$_{\text{posts-only Ask}}$} &
\yifan{83.69} & \yifan{94.02} & \yifan{92.42} & \yifan{71.51} & \yifan{85.41} \\
ROME$_{\text{w/o pretrain}}$     & 47.58 & 92.19 & 80.14 & 56.93 & 69.21 \\
\midrule
\textbf{ROME} & \textbf{92.24} & \textbf{95.90} & \textbf{97.45} & \textbf{81.32} & \textbf{91.73} \\
\bottomrule
\end{tabular}
% \end{table}
\vspace{-0.3cm}
\end{wraptable}

\noindent\textbf{Benefits of psychological knowledge.} We
further ablate each information source in
isolation to examine their individual contributions and complementarity. Specifically, using only questionnaire-grounded evidence while discarding post representations
(ROME$_{\text{w/o posts}}$) reduces the average Macro-F1
to 75.12. Conversely, using only post representations (ROME$_{\text{w/o evidence}}$) leads to an even
larger drop to 65.77. This pattern aligns with
our intuition: post-derived features capture
broad and expressive linguistic cues, whereas
questionnaire-grounded evidence provides semantically aligned and psychologically meaningful
signals. ROME performs best when both sources are jointly leveraged.

\noindent\yifan{\textbf{Benefits of combining posts and labels in role-play.} We also examine the contributions of posts and labels to Ask-stage response generation while keeping all other components fixed. Removing personality labels from the role-playing LLM's inputs (ROME$_{\text{posts-only Ask}}$) reduces the average Macro-F1 from 91.73 to 85.41. Conversely, retaining only personality labels (ROME$_{\text{label-only Ask}}$) leads to a larger drop to 68.42, only 2.65 points above removing the evidence pathway entirely. These results indicate that label expansion alone provides limited auxiliary supervision, while grounding role-play responses in user posts is important for effective question-level supervision.
Further evidence is provided in Appendix~\ref{app:response-diversity}.}

{\normalcolor
\noindent\textbf{Benefits of question-conditioned expert routing.}
We introduce two orthogonal design choices: (i) whether question information enters the predictor, and (ii) what the router conditions on.

\begin{wraptable}{r}{0.52\textwidth}
\centering
\caption{\yifan{Ablation study of the Answer module on the \textsc{Kaggle} dataset.}}
\label{tab:moe-ablation}
\small
\setlength{\tabcolsep}{4pt}
\renewcommand{\arraystretch}{0.9}
\begin{tabular}{lccccc}
\toprule
\textbf{\yifan{Methods}} &
\textbf{\yifan{I/E}} &
\textbf{\yifan{S/N}} &
\textbf{\yifan{T/F}} &
\textbf{\yifan{P/J}} &
\textbf{\yifan{Avg}} \\
\midrule

\yifan{ROME$_{\text{user-only MLP}}$}
& \yifan{45.15} & \yifan{91.96} & \yifan{80.27}
& \yifan{57.90} & \yifan{68.82} \\

\yifan{ROME$_{\text{q-cond. MLP}}$}
& \yifan{64.14} & \yifan{92.83} & \yifan{88.82}
& \yifan{66.76} & \yifan{78.14} \\

\yifan{ROME$_{\text{uniform MoE}}$}
& \yifan{82.58} & \yifan{93.16} & \yifan{93.30}
& \yifan{71.16} & \yifan{85.05} \\

\yifan{ROME$_{\text{user-cond. MoE}}$}
& \yifan{\underline{91.46}} & \yifan{93.46}
& \yifan{95.29} & \yifan{\underline{76.98}}
& \yifan{\underline{89.30}} \\

\midrule

\textbf{\yifan{ROME}}
& \textbf{\yifan{92.24}}
& \textbf{\yifan{95.90}}
& \textbf{\yifan{97.45}}
& \textbf{\yifan{81.32}}
& \textbf{\yifan{91.73}} \\

\bottomrule
\end{tabular}
\end{wraptable}

\par\noindent\textbf{(1) Question information improves predictor.}
Predicting all 60 items from user representation alone (ROME$_{\text{user-only MLP}}$) forces a single mapping to serve heterogeneous items and recovers only 3.05 points over removing the evidence pathway entirely. Conditioning the predictor on the item text and its construct indicator (ROME$_{\text{q-cond. MLP}}$) improves the average Macro-F1 from 68.82 to 78.14, confirming that different items require different mappings from user language to responses.

\par\noindent\textbf{(2) Question information improves router.}
We note that the uniform MoE, user-conditioned MoE, and full ROME variants have identical MoE structure, differing only in the router design. Replacing uniform mixing (ROME$_{\text{uniform MoE}}$) with user-conditioned routing (ROME$_{\text{user-cond. MoE}}$) improves the average Macro-F1 from 85.05 to 89.30, and additionally conditioning the router on question information (ROME) further improves it from 89.30 to 91.73. This isolates the contribution claimed by our method: user-dependent routing alone is insufficient, and the router needs to know which construct is being probed in order to select experts specialized for the corresponding psycholinguistic cues and associated linguistic cues.
\par}

\noindent\textbf{Benefits of model pretraining.} Finally, skipping the pretraining stage of the answer module (ROME$_{\text{w/o pretrain}}$) leads to a substantial performance reduction to $69.21$. This indicates that directly optimizing the full model without first stabilizing the question-level prediction task makes it difficult for ROME to fully leverage item-level supervision signals.

% Taken together, these ablation results indicate that question-level weighting, adaptive fusion of post and questionnaire evidence, and pretraining of the answer module are all essential for realizing the full performance gains of ROME.

% \begin{table}[t]
% \caption{Results of ablation study on \textsc{Kaggle} dataset (F1\%).}
% \label{tab:model-variants}
% \centering
% \normalsize
% \setlength{\tabcolsep}{5pt}
% \begin{tabular}{l|cccc|c}
% \hline
% \textbf{Methods} & \textbf{I/E} & \textbf{S/N} & \textbf{T/F} & \textbf{P/J} & \textbf{Avg} \\
% \hline
% ROME$_{\text{w/o q-weighting}}$  & \underline{90.13} & \underline{94.82} & \underline{96.36} & 72.55 & 88.47 \\
% ROME$_{\text{w/o gated fusion}}$ & 89.03 & 94.45 & \underline{96.36} & \underline{74.16} & \underline{88.50} \\
% ROME$_{\text{w/o posts}}$        & 86.56 & 71.21 & 94.18 & 48.51 & 75.12 \\
% ROME$_{\text{w/o evidence}}$     & 42.33 & 92.35 & 75.11 & 53.27 & 65.77 \\
% ROME$_{\text{w/o pretrain}}$     & 47.58 & 92.19 & 80.14 & 56.93 & 69.21 \\
% \hline
% \textbf{ROME} & \textbf{92.24} & \textbf{95.90} & \textbf{97.45} & \textbf{81.32} & \textbf{91.73} \\
% \hline
% \end{tabular}
% \end{table}

% \subsection{LLM-Based Inference Methods}
\subsection{Inference-Time LLM Methods}
\label{subsec:llm-inference}

To better understand different ways of using LLMs for personality detection, we compare three representative settings on the \textsc{Kaggle} test set, as summarized in Table~\ref{tab:llm_performance}: direct prompting with ChatGPT, recent inference-time LLM reasoning methods, and our ROME framework with different Ask-stage LLM backbones. Following \cite{hu2024tae}, we first consider two prompting configurations for \texttt{gpt-3.5-turbo} and \texttt{gpt-4o-2024-08-06} under zero-shot and 3-shot settings. We further include recent online inference methods, including \texttt{PsyCoT-3.5}, \texttt{HIPPD-4o}, and \texttt{EERPD-3.5}, all of which rely on inference-time LLM interactions and substantially heavier generation pipelines, making them fundamentally different from our setting.

% The results reveal a clear trend. Direct prompting baselines already benefit from stronger backbone LLMs, with \texttt{ChatGPT-4o} substantially outperforming \texttt{ChatGPT-3.5}, while 3-shot prompting brings only limited and inconsistent gains. 
Compared with direct prompting, online reasoning-based methods improve performance by introducing more structured inference procedures, such as multi-turn questionnaire prompting or explicit 
\begin{wraptable}{r}{0.52\textwidth}
\centering
\caption{Inference-Time LLM performance on the \textsc{Kaggle} dataset.}
\label{tab:llm_performance}
\small
\setlength{\tabcolsep}{4pt}
\renewcommand{\arraystretch}{0.9}
\begin{tabular}{lccccc}
\toprule
\textbf{Methods} & \textbf{I/E} & \textbf{S/N} & \textbf{T/F} & \textbf{P/J} & \textbf{Avg} \\
\midrule
ChatGPT-3.5                    & 65.86 & 51.69 & 78.60 & 63.93 & 65.02 \\
ChatGPT-3.5$_{\text{3-shot}}$  & 70.61 & 58.35 & 76.58 & 65.43 & 67.74 \\
ChatGPT-4o                     & 81.70 & 78.15 & 85.13 & 74.67 & 79.91 \\
ChatGPT-4o$_{\text{3-shot}}$   & 80.55 & 76.53 & 84.87 & 78.13 & 80.02 \\
\midrule
PsyCoT-3.5 & 66.56 & 61.70 & 74.80 & 57.83 & 65.22 \\
HIPPD-4o   & 73.61 & 77.25 & 84.98 & 79.33 & 78.79 \\
EERPD-3.5  & 86.63 & 90.59 & 89.15 & \textbf{82.12} & 87.12 \\
\midrule
\textbf{ROME-3.5} & \underline{88.32} & \underline{94.91} & \underline{95.86} & 73.76 & \underline{88.21} \\
\textbf{ROME-4o}  & \textbf{92.24} & \textbf{95.90} & \textbf{97.45} & \underline{81.32} & \textbf{91.73} \\
\bottomrule
\end{tabular}
% \end{table}
\vspace{-0.3cm}
\end{wraptable}
reasoning over psychologically relevant cues. 
% Among them, \texttt{EERPD-3.5} is particularly strong, achieving an average Macro-F1 of $87.12$ and even outperforming \texttt{HIPPD-4o} and all direct ChatGPT prompting variants. 
Nevertheless, both \texttt{ROME-3.5} and \texttt{ROME-4o} outperform all these alternatives in overall performance. In particular, \texttt{ROME-3.5} already surpasses the strongest online reasoning method \texttt{EERPD-3.5}, while \texttt{ROME-4o} achieves the best average Macro-F1 of $91.73$. This result suggests that the advantage of ROME does not merely come from using a stronger backbone LLM, but from its questionnaire-grounded Ask--Answer--Detect framework. Instead of directly relying on raw inference-time LLM outputs for each test instance, ROME transforms LLM-based reasoning into offline supervision by generating question-level answers during training. Importantly, because such generated answers can be noisy and vary in reliability across questions, ROME further calibrates this supervision by considering both the stability of generated responses and the diagnostic importance of questionnaire items. 

%  The model then learns to reproduce this psychologically grounded intermediate evidence and fuse it with post representations for final prediction. In this way, ROME provides a more explicit and better aligned mapping from free-form user language to abstract personality constructs.
% \begin{table}[t]

More importantly, unlike online inference-time methods such as \texttt{PsyCoT}, \texttt{HIPPD}, and \texttt{EERPD}, ROME uses the LLM only in the offline Ask stage during model training, while inference is performed entirely by a compact supervised model. This design yields not only better overall accuracy, but also substantially lower latency and monetary cost, making ROME more practical for deployment.

\subsection{Data Efficiency under Limited Supervision}
\label{subsec:data-efficiency}

% \begin{figure}[t]
\begin{wrapfigure}{r}{0.37\textwidth}
\vspace{-0.5cm}
  \centering
  \includegraphics[width=0.98\linewidth]{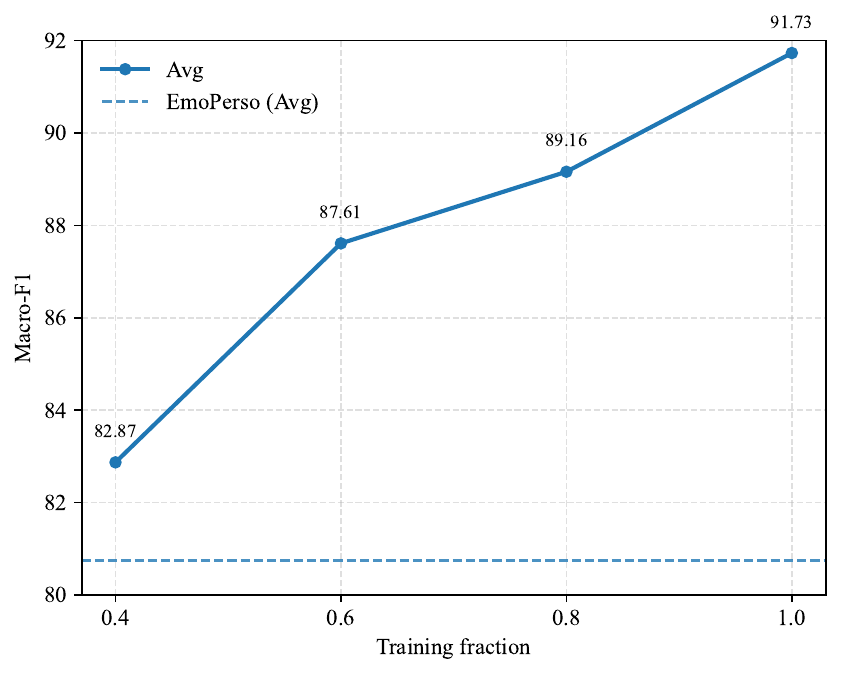}
  \caption{Effect of training data size on the \textsc{Kaggle} dataset.}
  \label{fig:abl_datafraction}
% \end{figure}
\vspace{-0.5cm}
\end{wrapfigure}

To examine how well ROME performs when labeled data are scarce, a situation frequently encountered in practice because personality annotation is costly and often subject to privacy concerns, we simulate limited-supervision settings by down-sampling the \textsc{Kaggle} training set while keeping the validation and test sets fixed. Concretely, we train the same ROME configuration
% (60-item questionnaire and 32 experts) 
using only $40\%$, $60\%$, $80\%$, or $100\%$ of the training data and report the resulting average Macro-F1 in Figure~\ref{fig:abl_datafraction}. As the available training data increases, performance improves smoothly from $82.87$ (40\%) to $87.61$ (60\%), $89.16$ (80\%), and $91.73$ with the full dataset, indicating that the model behaves stably across data scales and can effectively exploit additional supervision when it is available.

More importantly, ROME remains highly competitive even in the most data-poor regime. With only $40\%$ of the training data, our model ($82.87$) already  surpasses the best baseline method EmoPerso, whose average Macro-F1 under full-data training is $80.75$ (Table~\ref{tab:main_results}). 
% When trained on $60\%$ and $80\%$ of the data, ROME outperforms ETM by approximately $2.12$ and $4.59$ percentage points, respectively. 
These results highlight the data efficiency of the Ask--Answer--Detect framework: by leveraging standardized questionnaires and question-level supervision, ROME benefits from a much denser and more structured training than traditional ``posts $\rightarrow$ user vector $\rightarrow$ labels'' models, and thus maintains strong performance even when labeled users are very limited.

\subsection{\yifan{Additional Experimental Analyses}}
\label{subsec:additional-analysis}
\yifan{To complement the main experiments, we conduct further analyses in the Appendix. We first examine the effects of backbone LLMs and text encoders (Appendix~\ref{app:backbone}), as well as questionnaire length and the number of MoE experts (Appendix~\ref{app:param-analysis}). We further investigate the interpretability of ROME through question--post evidence chains, item-removal sensitivity, and expert activation patterns (Appendix~\ref{app:case-study}). In addition, we analyze the diversity of role-played questionnaire responses to examine whether they preserve user-specific variation beyond personality labels (Appendix~\ref{app:response-diversity}). Finally, we provide an additional comparison with EERPD on \textsc{Pandora} and evaluate the applicability of ROME to Big Five personality traits on the Essays dataset (Appendix~\ref{app:additional-experiments}). These experiments provide supplementary evidence on the robustness, interpretability, and broader applicability of ROME.}

\section{Conclusion}
In this paper, we propose a role-play–enhanced, question-conditioned Mixture-of-Experts framework for personality detection that replaces the prevailing ``posts $\rightarrow$ user vector $\rightarrow$ labels'' paradigm with an Ask--Answer--Detect pipeline. By leveraging an LLM-based role-play agent to generate question-level answers and treating these answers as questionnaire-grounded evidence, our framework injects structured psychological knowledge into the modeling process and provides an explicit question-level reasoning chain from raw text to personality labels. Extensive experiments on two real-world datasets demonstrate substantial and consistent gains, with relative Macro-F1 improvements of up to 13.60\% and 19.72\% over state-of-the-art baselines.

% \begin{figure}[h]
%   \centering
%   \includegraphics[width=0.90\linewidth]{fig/expert_dim_heatmap}
%   \caption{Expert activation analysis on \textsc{Kaggle} dataset.}
%   \label{fig:expert-dim-heatmap}
% \end{figure}

% together with fine-grained interpretability at the level of questions, experts, and case studies. In future work, we plan to combine the questionnaire structure with external knowledge graphs grounded in psychological and personality theories, further enriching the questionnaire-grounded evidence space and supporting more nuanced, theory-informed reasoning about personality from online text.

% \newpage

\begin{ack}
\yifan{This work was supported by the Guangzhou-HKUST(GZ) Joint
Funding Program (No.~2024A03J0630).} The authors declare no competing interests.
\end{ack}
\bibliographystyle{plainnat}
\bibliography{references_mbti}
\clearpage
\appendix

\section{Related Work}

\subsection{Personality Detection}
Early evidence shows that digital traces and language use carry
robust signals of human personality \cite{Schwartz2013PLOS,Youyou2015PNAS}. Classical approaches
therefore relied on handcrafted or lexicon-based features (\textit{e.g.}, LIWC)
to engineer user representations for personality classification \cite{Mairesse2007,Pennebaker2015LIWC}.
With the advent of deep learning, the community has shifted
toward neural encoders that aggregate multiple posts into a unified user vector automatically, typically built upon pretrained language models such as BERT \cite{kenton2019bert}, with strategies ranging from simple concatenation to hierarchical pooling \cite{jiang2020automatic,keh2019myers}. To better capture dependencies across posts, subsequent studies further introduce cross-document modeling and graph-based architectures (\textit{e.g.}, multi-document transformers and dynamic graph convolution) to enhance user-level semantic representations \cite{yang2021mdt,yang2023ddgcn,zhu2022contrastive}. Some recent extensions further enhance user representations by incorporating psychological guidance or multi-view modeling. For example, \cite{yang2021pqnet} leverages psychological domain knowledge to extract personality-relevant cues from user posts, while \cite{zhu2024mvp} models users from multiple perspectives using consistency regularization.

More recently, LLM-augmented pipelines have advanced the frontier by leveraging LLMs' strong generation and embedding capacities. For instance, TAE \cite{hu2024tae} leverages LLMs to generate multi-perspective analyses of posts and labels within contrastive learning frameworks; ETM \cite{bi2025etm} strengthens user vectors with LLM-derived long-text embeddings and label-aware alignment; LL4G \cite{shen2025ll4g} incorporates LLM-derived semantic embeddings to facilitate graph construction for better graph-based user representation learning; and EmoPerso \cite{shen2025emoperso} introduces LLM-based generative mechanisms for conditioned paraphrasing and contextual feature completion to enrich emotion-aware user representations. Despite these advances, the prevailing modeling paradigm remains ``posts $\rightarrow$ user vector $\rightarrow$ labels'', which learns text-to-personality mappings only from coarse-grained labels, offering limited interpretability and intermediate supervision. 

In parallel, some recent studies further explore inference-time LLM reasoning for personality detection, including multi-turn questionnaire prompting \cite{yang2023psycot}, retrieval-augmented few-shot inference \cite{li2025eerpd}, and online hierarchical LLM-based reasoning \cite{chen2025hippd}. However, these approaches require online interaction with LLMs and incur substantial higher inference costs, making them fundamentally different from our setting. 

\subsection{Role-play Prompting}
Role-play has emerged as a popular prompt-based paradigm for conditioning LLMs on personas, goals, and contextual constraints across both simulation and evaluation settings, enabling coherent agent behaviors and goal-directed reasoning in open-ended environments \citep{li2023camel,park2023generativeagents}. Recent studies have systematized the use of role-play in LLMs by introducing benchmarks to quantify role adherence \cite{wang2024rolellm} and domain-specific frameworks that co-design standardized, faithful interactions such as clinician-guided virtual patients and character-grounded agents in text-based environments \cite{louie2024roleplaydoh,wang2025characterbox}. These developments, further synthesized in a recent survey \cite{tseng2024roleplay}, collectively underscore the growing interest in aligning LLM behavior with contextual roles. In this paper, we do not employ role-play for open-ended dialogue generation. Instead, we elicit question-level answers to validated psychometric questionnaires from users’ posts, providing psychologically grounded intermediate supervision for personality detection. To the best of our knowledge, this is the first work to harness LLM's role-play capacity to construct an explicit questionnaire-grounded evidence space for this task, tightly integrated with our question-conditioned Mixture-of-Experts architecture that learns to answer questionnaire items before final personality classification.

\subsection{Mixture-of-Experts (MoE)}
Mixture-of-Experts (MoE) realizes \emph{conditional computation} by routing each input to a small subset of experts, thereby achieving scalable model capacity through sparse activation \cite{shazeer2017moe,lepikhin2020gshard,fedus2021switch,du2022glam}. Beyond sheer scale, task- and signal-aware variants encourage functional specialization. For instance, multi-gate MoE architectures have been widely applied in multi-task learning scenarios \cite{ma2018mmoe}. More recent systems further develop different routing mechanisms that address various aspects, including improving routing stability, load balance, and cross-task transferability \cite{jiang2024mixtral,liu2024deepseekv2,zhao2024hypermoe,qiu2024rmoe}. These advances collectively show that what conditions the router matters: conditioning on the appropriate structure enables experts to capture distinct facets of the problem. Unlike prior work that conditions primarily on token- or task-level signals to improve efficiency or general accuracy, we repurpose the MoE framework for question-conditioned specialization: standardized questionnaire signals drive routing and supervision at the item level, producing intermediate, psychologically grounded evidence that is subsequently aggregated for personality inference. This reframes the role of the MoE from scaling the encoder to structuring the supervision and providing an interpretable bridge for the under-specified text-to-personality mapping emphasized in our introduction.

\section{Implementation Details}
\label{app:imple}
\begin{table}[t]
\centering
\caption{Statistics of the two datasets.}
\label{tab:dataset-stats}
\normalsize
\setlength{\tabcolsep}{7.0pt}
\renewcommand{\arraystretch}{1.05}
\begin{tabular}{lcccc}
\toprule
\textbf{Dataset} & \textbf{Types} & \textbf{Train} & \textbf{Validation} & \textbf{Test} \\
\midrule
\multirow{4}{*}{Kaggle}
& I/E & 4032/1173 & 1330/405 & 1314/421 \\
& S/N & 724/4481 & 230/1505 & 243/1492 \\
& T/F & 2388/2817 & 802/933 & 791/944 \\
& P/J & 3160/2045 & 1007/728 & 1074/661 \\
\midrule
\multirow{4}{*}{Pandora}
& I/E & 4314/1126 & 1425/388 & 1403/411 \\
& S/N & 621/4819 & 202/1611 & 205/1609 \\
& T/F & 3527/1913 & 1160/653 & 1164/650 \\
& P/J & 3211/2229 & 1064/749 & 1035/779 \\
\bottomrule
\end{tabular}
\end{table}

\subsection{Question-to-Construct Assignment}
To support question-conditioned routing in the Answer stage, each questionnaire item is associated with a fixed MBTI construct indicator, \textit{i.e.}, one of the four MBTI dimensions: I/E, S/N, T/F, or P/J. We obtain this assignment once before training by prompting an LLM to identify the primary MBTI dimension measured by each questionnaire item, and then keep the resulting mapping fixed throughout training and evaluation. The prompt used for this assignment is shown in Figure~\ref{fig:q2c_prompt}. The resulting fixed mapping from questionnaire items to MBTI dimensions is reported in Table~\ref{tab:q2dim}.

\begin{figure}[t!]
\centering
\resizebox{0.9\linewidth}{!}{
\begin{tcolorbox}[
    colback=gray!4,
    colframe=gray!60,
    title=Prompt for Question-to-Construct Assignment,
    width=1.0\linewidth,
    before upper={
        \footnotesize
        \linespread{1.0}\selectfont
        \setlength{\parskip}{2pt}
    }
]
\textbf{Role and Task:}\\
You are a research assistant in computational psychology and psychometrics. Your task is to assign each questionnaire item to exactly one primary MBTI dimension.

\textbf{MBTI Dimensions:}
\begin{itemize}[leftmargin=1.6em,itemsep=1pt,topsep=2pt]
    \item I/E = Introversion vs.\ Extraversion
    \item S/N = Sensing vs.\ Intuition
    \item T/F = Thinking vs.\ Feeling
    \item P/J = Perceiving vs.\ Judging
\end{itemize}

\textbf{Instructions:}
\begin{enumerate}[leftmargin=1.8em,itemsep=1pt,topsep=2pt]
    \item For each item, choose exactly one primary dimension from: I/E, S/N, T/F, P/J.
    \item Assign the dimension that best matches the main psychological construct measured by the item.
    \item If an item seems ambiguous, still choose the single closest dimension.
    \item Do not provide explanations, rationales, confidence scores, or any extra text.
\end{enumerate}

\textbf{Output Format:}\\
Return exactly one line per item in the following format: \\
\texttt{Q<item\_number>: <dimension>}

\textbf{Example:}
\begin{itemize}[leftmargin=1.6em,itemsep=1pt,topsep=2pt]
    \item \texttt{Q1: I/E}
    \item \texttt{Q2: S/N}
    \item \texttt{Q3: T/F}
\end{itemize}

\textbf{Items to Classify:}\\
\texttt{[Paste the 60 questionnaire items here]}
\end{tcolorbox}
}
\caption{Prompt used for question-to-construct assignment.}
\label{fig:q2c_prompt}
\end{figure}

\begin{table}[h]
\centering
\small
\caption{Fixed mapping from questionnaire items to MBTI dimensions.}
\label{tab:q2dim}
\setlength{\tabcolsep}{8pt}
\renewcommand{\arraystretch}{1.05}
\begin{tabular}{ll}
\toprule
\textbf{Question IDs} & \textbf{Assigned dimensions} \\
\midrule
Q1--Q10   & I/E, S/N, T/F, P/J, T/F, I/E, P/J, T/F, P/J, T/F \\
Q11--Q20  & I/E, S/N, T/F, P/J, T/F, I/E, S/N, T/F, S/N, T/F \\
Q21--Q30  & I/E, S/N, T/F, P/J, T/F, I/E, T/F, T/F, P/J, T/F \\
Q31--Q40  & I/E, S/N, T/F, P/J, P/J, I/E, S/N, T/F, P/J, T/F \\
Q41--Q50  & I/E, S/N, I/E, P/J, T/F, S/N, T/F, T/F, P/J, T/F \\
Q51--Q60  & I/E, S/N, I/E, T/F, T/F, P/J, S/N, T/F, P/J, T/F \\
\bottomrule
\end{tabular}
\end{table}

\subsection{Training Procedure and Computational Setup}
\yifan{Before role-play generation, we first perform the user-level 60/20/20 train/validation/test split. All role-play-based questionnaire responses in the Ask stage are generated exclusively for users in the training partition; no such responses are generated for validation or test users. The reliability score $q_{\mathrm{unc},i}$, importance score $q_{\mathrm{imp},i}$, and resulting question weight $w_i$ are likewise estimated using only the training partition.}

In the Ask stage, we employ ChatGPT (\texttt{gpt-4o-2024-08-06}) as a role-playing LLM to generate question-level answers to the MBTI-style questionnaire from user posts. The question-conditioned Mixture-of-Experts (MoE) module in the Answer stage consists of 32 experts, each instantiated as a lightweight MLP with a hidden dimension of 1024. In the first stage, we pretrain the Answer module with the regression loss $\mathcal{L}_q$ using a learning rate of $5\times 10^{-4}$, a mini-batch size of 64, and 120 training epochs. In the second stage, we jointly optimize the entire model with the multi-task objective $\mathcal{L} = \lambda_q \mathcal{L}_q + \lambda_{\mathrm{cls}} \mathcal{L}_{\mathrm{cls}}$, where $\lambda_q = 1$ and $\lambda_{\mathrm{cls}} = 0.05$, using a learning rate of $1\times 10^{-4}$ and a mini-batch size of 32. For the main results of ROME, we repeat training and evaluation over five times with different random seeds and report the mean and standard deviation of Macro-F1.

All experiments are conducted on a single NVIDIA RTX 5880 Ada GPU. The Ask stage relies on the ChatGPT API (\texttt{gpt-4o-2024-08-06}) only during data preprocessing, while the Answer stage is trained locally using PyTorch. The full training pipeline completes within approximately 1 hour, indicating the efficiency and practicality of the proposed framework.

\section{Effect of Backbone LLMs and Encoders}
\label{app:backbone}

\begin{figure}[t]
  \centering
  \begin{subfigure}[t]{0.5\linewidth}
    \centering
    \includegraphics[height=3.8cm,keepaspectratio]{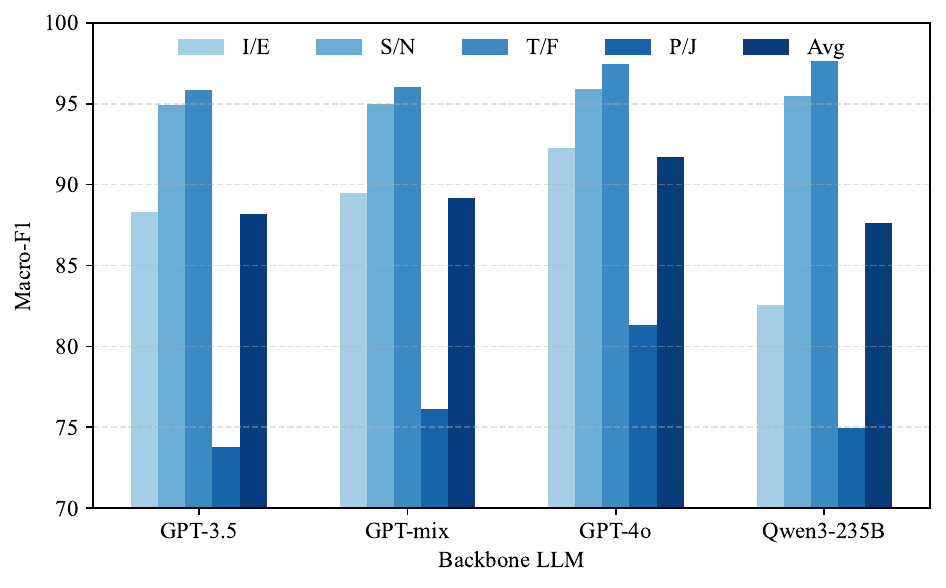}
    \caption{Backbone LLM comparison.}
    \label{fig:backbone-llm}
  \end{subfigure}
  \hfill
  \begin{subfigure}[t]{0.49\linewidth}
    \centering
    \includegraphics[height=3.8cm,keepaspectratio]{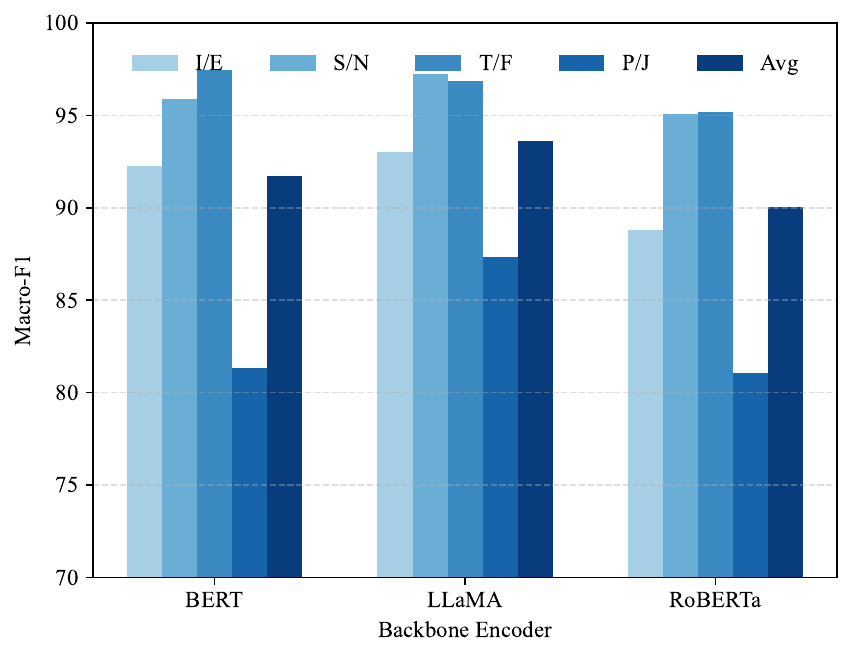}
    \caption{Backbone encoder comparison.}
    \label{fig:backbone-encoder}
  \end{subfigure}
  \caption{Effect of backbone LLMs and text encoders on \textsc{Kaggle} dataset.}
  \label{fig:backbone-half-left}
\end{figure}

To analyze the effect of backbone LLM choice, we vary the LLM deployed as the agent in the \emph{Ask} stage while holding all other components of ROME fixed. Concretely, we consider Qwen3-235B, GPT-3.5, GPT-4o, and a simple ensemble variant (GPT-mix) that averages the question-level answers generated by GPT-3.5 and GPT-4o before training the downstream modules. As shown in Figure~\ref{fig:backbone-llm}, all four settings yield strong results, but their average Macro-F1s follow a clear ordering: Qwen3-235B attains $87.65$, GPT-3.5 attains $88.21$, GPT-mix improves to $89.16$, and GPT-4o achieves the best performance at $91.73$. 
% The advantage of GPT-4o is especially visible on the more challenging I/E and P/J dimensions, where the weaker model degrades more noticeably. 
This pattern is consistent with the view that the \emph{quality of the role-play supervision} matters: GPT-4o produces more faithful, less noisy questionnaire answers during Ask stage, which in turn provide a reliable training signal for the Answer and Detect stages. Even so, the Qwen3-235B configuration already significantly exceeds the best baseline method EmoPerso (Table~\ref{tab:main_results}, $80.75$), indicating that ROME remains effective under more economical LLM choices.

We then examine the impact of the backbone text encoder used to represent posts and questionnaire items. Figure~\ref{fig:backbone-encoder} reports results when using BERT-base-uncased (the default configuration in Table~\ref{tab:main_results}), Meta-Llama-3-8B-Instruct as a frozen encoder, and RoBERTa-base, with the Ask-stage LLM and all other components unchanged. BERT already supports a strong average Macro-F1 of $91.73$, while replacing it with LLaMA further raises the score to $93.61$.
% with consistent gains on all four dimensions and a particularly large improvement on P/J (from $77.95\%$ to $85.91\%$).  RoBERTa remains competitive but lags behind BERT with an average Macro-F1 of $85.93\%$, mainly due to weaker performance on I/E and P/J prediction. 
Overall, these findings indicate that ROME is relatively robust across encoder choices—both BERT and RoBERTa configurations already outperform the best baseline methods—while the LLaMA-based variant further benefits from its stronger capacity to encode heterogeneous, long-range semantics in user posts and questionnaire items. 

\section{Parameters Analysis}
\label{app:param-analysis}
\subsection{Effect of questionnaire length}
We first investigate how many questionnaire items are required for ROME to perform well. As shown in Figure~\ref{fig:abl_questions}, we randomly subsample a fixed number of items from the 60-item inventory (12, 24, 36, 48, or 60) and retrain the model on \textsc{Kaggle} under each setting. The average Macro-F1 increases steadily from 84.16 with 12 items to 91.73 with all 60 items, indicating that additional evidence systematically strengthens the psychology-grounded reasoning process and yields more accurate personality predictions. 
% At the personality dimension level, S/N and T/F remain strong across all settings, while P/J benefits most from longer questionnaires, showing the largest performance gain as more items are included.
Despite this clear upward trend, ROME remains highly competitive even with substantially shortened questionnaires. With 12 and 24 randomly sampled items, the model attains average Macro-F1s of 84.16 and 86.33, respectively, already outperforming the best baseline method EmoPerso (80.75, Table~\ref{tab:main_results}) by relative margins of $4.22\%$ and $6.91\%$. This demonstrates that the Ask--Answer--Detect framework can extract robust, psychology-aligned evidence even from incomplete questionnaires. 

\subsection{Effect of the number of experts}
\begin{figure}[t]
  \centering
  \begin{subfigure}[t]{0.42\textwidth}
    \centering
    \includegraphics[width=\linewidth]{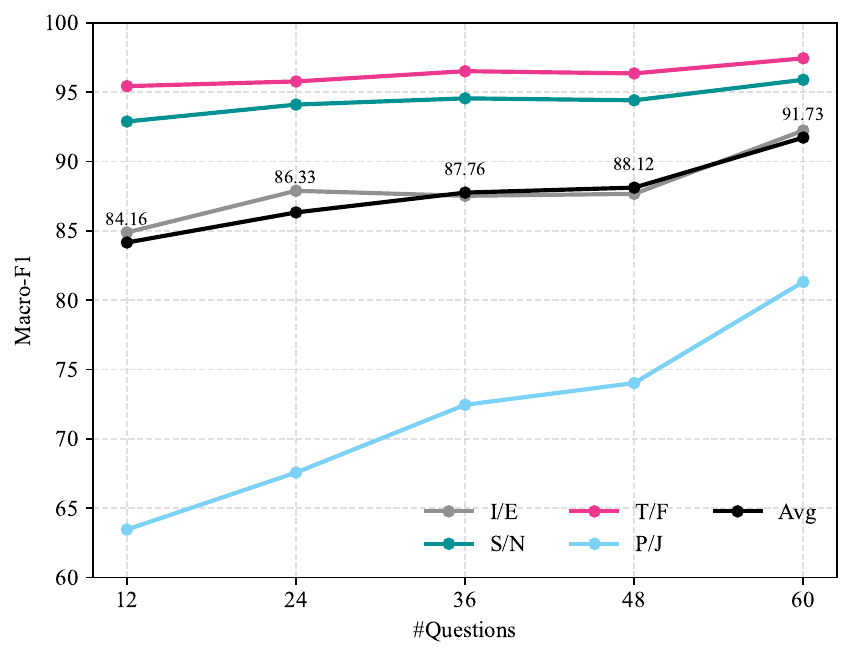}
    \caption{Effect of questionnaire length}
    \label{fig:abl_questions}
  \end{subfigure}
  \hfill
  \begin{subfigure}[t]{0.42\textwidth}
    \centering
    \includegraphics[width=\linewidth]{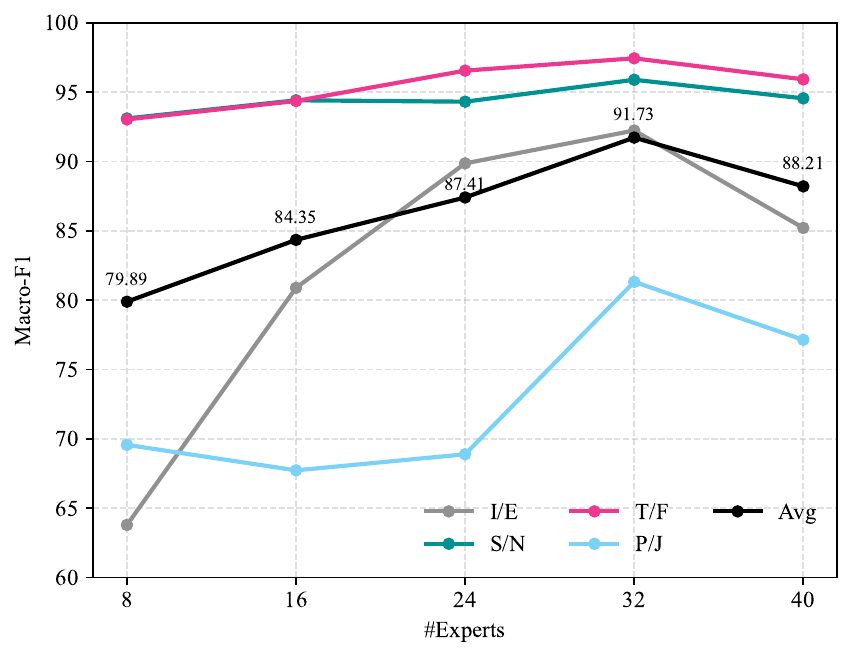}
    \caption{Effect of the number of experts}
    \label{fig:abl_experts}
  \end{subfigure}

  \caption{Effect of questionnaire length and the number of experts on the \textsc{Kaggle} dataset.}
  \label{fig:abl_questions_experts}
\end{figure}

We next study how the capacity of the question-conditioned MoE affects performance by varying the number of experts while keeping all other settings fixed. As shown in Figure~\ref{fig:abl_experts}, increasing the number of experts from 8 to 32 steadily improves the average Macro-F1 on \textsc{Kaggle} dataset from $79.89$ to $91.73$, whereas further enlarging the expert pool to 40 yields a slight drop to $88.21$. This pattern suggests that a small expert set provides insufficient capacity to capture the heterogeneous psycholinguistic cues associated with different questionnaire items and personality dimensions, while a moderately sized expert pool enables more effective specialization. Beyond this regime, however, the additional experts introduce extra complexity and routing difficulty without translating into further gains. 
% Overall, using 32 experts strikes a favorable balance between model capacity and stability, and we adopt this configuration as the default setting in all main experiments.

% At the dimension level, we observe that the I/E and P/J dimensions benefit most from increasing the number of experts, with substantial improvements as the expert pool grows from 8 to 32. In contrast, the S/N and T/F dimensions remain strong and relatively stable once a modest number of experts is available. These results indicate that different personality dimensions exhibit different levels of heterogeneity in their linguistic manifestations, and that a moderate expert capacity is sufficient for ROME to realize the advantages of question-conditioned specialization across all four dimensions.

% \begin{figure}[t]
%   \centering
%   \includegraphics[width=0.90\linewidth]{fig/abl_experts}
%   \caption{Effect of the number of experts on \textsc{Kaggle} dataset  (F1\%).}
%   \label{fig:abl_experts}
% \end{figure}

\section{Case Study}
\label{app:case-study}
To understand how ROME integrates questionnaire items and textual evidence to form personality predictions, we conduct a three-part interpretability analysis that examines the model’s behavior. 
% We begin with a fine-grained case study that illustrates how individual questionnaire items align with semantically relevant post snippets to support model decisions. We then move to a quantitative sensitivity analysis that measures the functional contribution of each item by perturbing the evidence available at inference time. 
% Finally, we examine the routing behavior of the question-conditioned MOE to reveal how ROME organizes its internal capacity in a manner that reflects the structure of the MBTI psychometric questionnaires.

\subsection{Question–Post Evidence Chains}

To make the Ask--Answer--Detect pipeline more concrete, we present a qualitative case study for a randomly selected test user whose true type is \texttt{ISTJ} and for whom ROME correctly predicts all four MBTI dimensions. As shown in Table~\ref{tab:case-study-intj}, for each dimension we select a handful of questionnaire items with the largest impact on the final decision, measured by the change in the predicted logit when the item is removed from the evidence aggregation. We then retrieve representative snippets from the user’s posts that GPT-4o, in a post-hoc analysis, identifies as semantically aligned with these high-impact items,  with key phrases highlighted for readability.

Across dimensions, the question--post pairs reveal coherent, psychologically plausible evidence patterns. For example, on the I/E axis, the most influential items are supported by snippets describing repeatedly turning down social invitations, a strong need for personal space, and discomfort with close contact or crowded environments, all pointing toward low social needs and a clear inclination for solitude. 
% For S/N, the selected items and snippets jointly show a mixture of concrete, detail-oriented advice (e.g., following others’ r\'esum\'e styles) and reflections on perspectives and interpretations, suggesting that the model integrates both experiential and interpretive cues when positioning this user. On T/F, high-weight items are linked to statements about questioning the reliability of information, reading up on relevant material, and evaluating situations through reasoning, while occasional expressions of guilt and concern for others appear as weaker, counterbalancing F-oriented signals. 
And for P/J, the evidence clusters around perfectionistic attitudes toward assignments, ongoing self-evaluation, and a strong drive to accomplish personally meaningful goals, reflecting a structured, planful, and self-disciplined J-oriented pattern. 

Overall, this case study demonstrates that ROME’s personality predictions are supported by a layered chain of question-level diagnostics and post-level linguistic cues.

\newcommand{\hlc}[2]{%
  {\setlength{\fboxsep}{0pt}
   \colorbox{#1}{#2}}%
}
\begin{table*}[t]
\caption{Illustrative question--post evidence for a representative \texttt{ISTJ} user.}
\label{tab:case-study-intj}
\small
\setlength{\tabcolsep}{4pt}
\centering
\resizebox{0.85\textwidth}{!}{%
\begin{tabular}{l|l|p{0.3\textwidth}|c||c|p{0.4\textwidth}}
\hline
\textbf{Dim.} & \textbf{Q\#} & \textbf{Question} & \textbf{Direction} & $\boldsymbol{\Delta}$\textbf{logit} & \textbf{Text evidence (snippet)} \\
\hline\hline
I/E & Q21 & You enjoy solitary hobbies or activities more than group ones. & I & $-3.19$ & ``...Feeling cozy in over sized sweaters, \hlc{eviA}{reading a good book} in \hlc{eviA}{a warm bed}...'' \\
\hline
% I/E & Q43 & You can easily connect with people you have just met. & I & $-0.84$  & ``I find myself \hlc{eviA}{turning down} anyone who asks me out or shows interest in me, even if we get along well...'' \\
% \hline
I/E & Q41 & You avoid making phone calls. & I & $-0.62$  & ``The very thought of having my \hlc{eviA}{personal space} violated in that manner by a complete stranger is terrifying to me.'' \\
\hline\hline
S/N & Q2  & Complex and novel ideas excite you more than simple and straightforward ones. & S & $-1.71$  & ``Usually, I don't really like \hlc{eviB}{discussing topics} like these but the posts here are so relatable.'' \\
\hline
S/N & Q42 & You enjoy exploring unfamiliar ideas and viewpoints. & N & $+0.83$ & ``That's an \hlc{eviB}{interesting perspective} -- that we shouldn't hold others' imperfections against them.'' \\
% \hline
% S/N & Q57 & You prefer tasks that require you to come up with creative solutions rather than follow concrete steps. & S & $-0.45$ & ``Try comparing your resume with the people you know who've gotten hired. Try to \hlc{eviB}{follow their style}...'' \\
\hline\hline
T/F & Q3  & You usually feel more persuaded by what resonates emotionally with you than by factual arguments. & T & $-1.54$ & ``That map isn't \hlc{eviC}{very reliable}. I know INFPs who live outside of Europe/Americas and there isn't a single dot...'' \\
\hline
% T/F & Q58 & You are more likely to rely on emotional intuition than logical reasoning when making a choice. & T & $-1.13$ & ``I've recently \hlc{eviC}{read up on} body language in an effort to improve my social skills. I seem to be able to read people well.'' \\
% \hline
T/F & Q15 & You rarely worry about whether you make a good impression on people you meet. & F & $+1.05$ & ``As a child and adolescent, I would lay awake for nights in a row with my stomach knotting and turning \hlc{eviC}{out of guilt}, even if the person said it didn't bother them.'' \\
\hline\hline
P/J & Q4  & Your living and working spaces are clean and organized. & J & $+1.02$ & ``If an assignment, essay, or other work was not perfect in my eyes, I really \hlc{eviD}{struggled just handing it in}.'' \\
\hline
P/J & Q34 & You find it challenging to maintain a consistent work or study schedule. & J & $+1.00$ & ``Lately, I have been \hlc{eviD}{evaluating my behaviors} and life and have concluded that I \hlc{eviD}{need to step}...'' \\
% \hline
% P/J & Q44 & If your plans are interrupted, your top priority is to get back on track as soon as possible. & J & $+0.51$ & ``I'd say that wanting to \hlc{eviD}{accomplish my goals} because of what is important to me keeps me going.'' \\
\hline\hline
\end{tabular}
}

\end{table*}

\subsection{Sensitivity of Item Removal}
To evaluate the functional role of individual questionnaire items in ROME’s inference process and verify whether the learned question weights ($w_i$ in Equation (\ref{eq:triple-prior})) capture truly diagnostic evidence, we conduct a sensitivity analysis based on item removal where the trained model is kept fixed and only the questionnaire evidence is altered. For each user and each MBTI dimension, we remove exactly one item from the weighted question summary in three ways: deleting the item with the highest learned weight (ROME$_{\text{drop-max-question}}$), deleting the item with the lowest weight (ROME$_{\text{drop-min-question}}$), or randomly deleting one item (ROME$_{\text{drop-rand-question}}$). As shown in Table~\ref{tab:inference-ablation}, removing the highest-weight item leads to the largest degradation, reducing the average Macro-F1 from $91.73$ to $89.39$, 
% with the S/N dimension dropping by more than $10$ points. Random deletion produces a milder decline ($89.12\%$), 
while removing the lowest-weight item leaves the overall performance almost unchanged. This pattern indicates that the learned question weights meaningfully reflect the diagnostic value of individual items: high-weight questions carry indispensable evidence for the final decision, while low-weight questions play a marginal role.
% , and the model remains reasonably robust under mild evidence loss.

\begin{table}[h]
\caption{Sensitivity of item removal on \textsc{Kaggle} dataset.}
\label{tab:inference-ablation}
\centering
\normalsize
\setlength{\tabcolsep}{5pt}
\begin{tabular}{lccccc}
\hline
\textbf{Methods} & \textbf{I/E} & \textbf{S/N} & \textbf{T/F} & \textbf{P/J} & \textbf{Avg} \\
\hline
ROME$_{\text{drop-max-question}}$   & 92.06 & 88.16 & 96.33 & 81.02 & 89.39 \\
ROME$_{\text{drop-rand-question}}$  & 92.01 & 95.08 & 97.21 & 81.09 & 91.35 \\
ROME$_{\text{drop-min-question}}$   & \underline{92.16} & \underline{95.81} & \textbf{97.45}& \textbf{81.43}& \underline{91.71} \\
\hline
\textbf{ROME} & \textbf{92.24} & \textbf{95.90} & \textbf{97.45} & \underline{81.32} & \textbf{91.73} \\
\hline
\end{tabular}
\end{table}

\begin{figure}[h]
  \centering
  \includegraphics[width=0.6\linewidth]{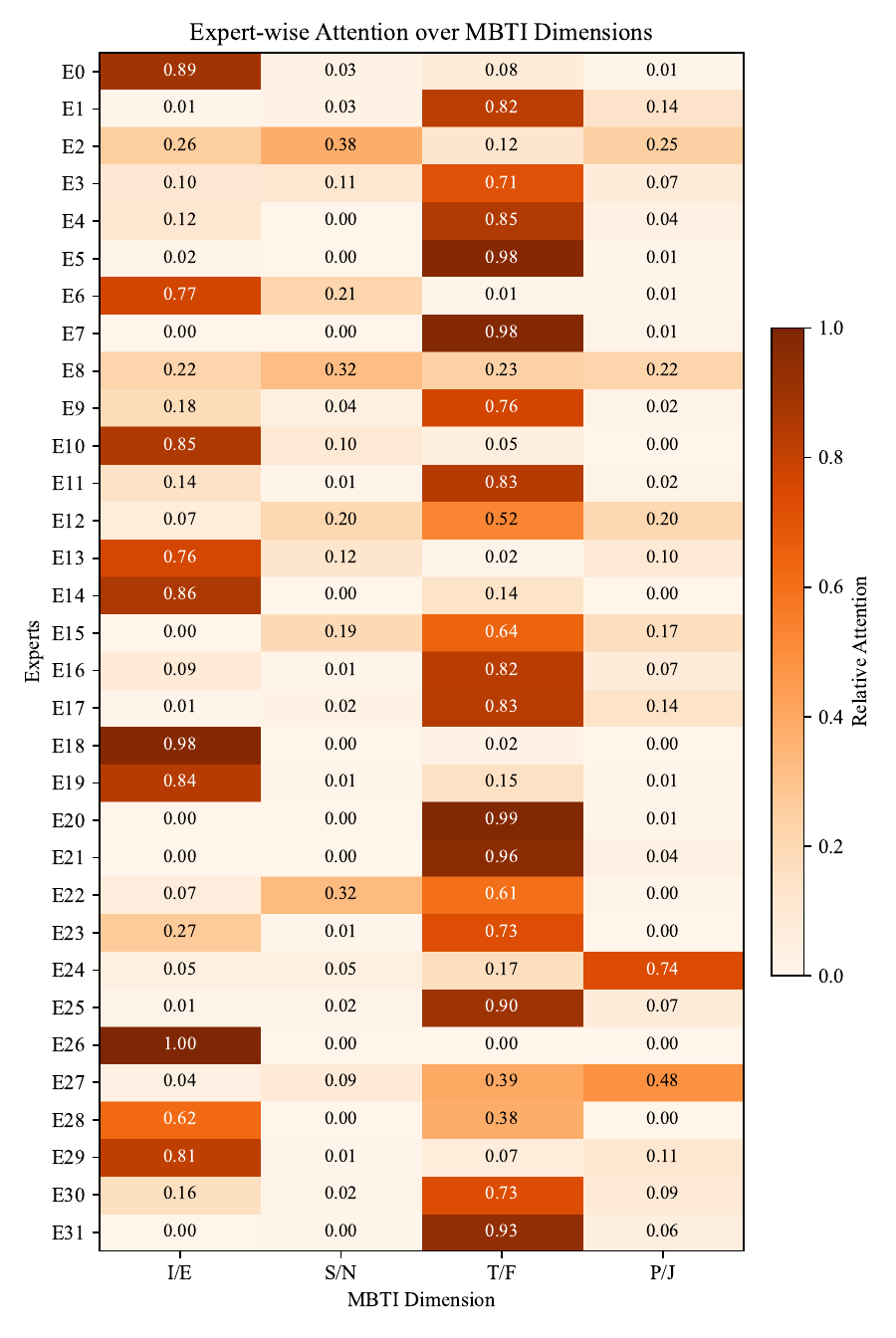}
  \caption{Expert activation analysis on \textsc{Kaggle} dataset.}
  \label{fig:expert-dim-heatmap}
\end{figure}

\subsection{Expert Activation Analysis: Internal Specialization in the MoE}
To further interpret our model, we analyze how the question-conditioned MoE allocates expert capacity across personality dimensions. For each user and each questionnaire item, we record the soft routing probabilities assigned to the experts and aggregate them over all users and all items associated with each MBTI dimension. Normalizing these aggregated values yields a $32\times 4$ matrix reflecting the \emph{relative attention} each expert devotes to the four MBTI dimensions (\textit{i.e.}, I/E, S/N, T/F, and P/J).

As visualized in Figure~\ref{fig:expert-dim-heatmap}, many experts exhibit a clear dominant dimension, concentrating the majority of their attention on one of the four MBTI axes, while others display more mixed routing patterns across dimensions. This specialization suggests that the MoE develops a meaningful division of labor, with different experts focusing on distinct psychological constructs and personality dimensions, while a small group of shared experts capturing cross-dimensional knowledge.

% \subsubsection{Expert Activation Analysis: Internal Specialization in the MoE}
% To further interpret our model, we analyze how the question-conditioned MoE allocates expert capacity across personality dimensions. For each user and each questionnaire item, we record the soft routing probabilities assigned to the experts and aggregate them over all users and all items associated with each MBTI dimension. Normalizing these aggregated values yields a $32\times 4$ matrix reflecting the \emph{relative attention} each expert devotes to the four MBTI dimensions (\textit{i.e.}, I/E, S/N, T/F, and P/J).

% As visualized in Figure~\ref{fig:expert-dim-heatmap}, many experts exhibit a clear dominant dimension, concentrating the majority of their attention on one of the four MBTI axes, while others display more mixed routing patterns across dimensions. This specialization suggests that the MoE develops a meaningful division of labor, with different experts focusing on distinct psychological constructs and personality dimensions, while a small group of shared experts capturing cross-dimensional knowledge.

% Taken together with the item-level case study, this expert-level analysis suggests that ROME does not merely increase model capacity, but organizes its experts in a manner that reflects the psychometric structure of the questionnaire, thereby providing an additional layer of interpretability for how question-level evidence is routed and aggregated.
\begingroup
\ifshowchanges
\color{yifancolor}
\fi

\Needspace{5\baselineskip}
\begingroup
\ifshowchanges\color{yifancolor}\fi

\subsection{Diversity of Role-Played Questionnaire Responses}
\label{app:response-diversity}

Complementing the Ask-stage input ablations in Section~\ref{subsec:ablation}, we further examine whether the role-played questionnaire responses preserve user-specific variation or collapse into similar response patterns for users sharing the same personality label. Specifically, we compute pairwise mean absolute error (MAE) between the 60-dimensional response vectors under three comparison settings: repeated generations for the same user, different users sharing the same MBTI type, and different users with different MBTI types.

\begin{table}[H]
\centering
\normalcolor
\caption{Pairwise MAE between role-played questionnaire-response vectors.}
\label{tab:response-diversity}
\small
\setlength{\tabcolsep}{5pt}
\renewcommand{\arraystretch}{0.9}
\begin{tabular}{lcc}
\toprule
\textbf{Comparison Setting} & \textbf{Mean MAE} & \textbf{Median MAE} \\
\midrule
Same user, repeated generations & 0.282 & 0.267 \\
Same MBTI type, different users & 0.529 & 0.477 \\
Different MBTI types, different users & 1.912 & 1.843 \\
\bottomrule
\end{tabular}
\end{table}

As shown in Table~\ref{tab:response-diversity}, repeated generations for the same user yield a mean pairwise MAE of 0.282, reflecting variation introduced by stochastic generation. The distance increases to 0.529 for different users sharing the same MBTI type, indicating that users with the same label still exhibit distinct response profiles beyond within-user generation variability. The distance further increases to 1.912 for users with different MBTI types, showing that the generated responses also preserve broader trait-level distinctions. Overall, the generated responses do not collapse into a single response template determined solely by the personality label.

Table~\ref{tab:same-type-example} provides an illustrative item-level example. Although both users are labeled ISTP, their responses to Q37, ``You are drawn to various forms of creative expression,'' are consistently opposite across five generations. The corresponding post excerpts also express contrasting attitudes toward creative activities.

\begin{table}[H]
\centering
\normalcolor
\caption{Illustrative role-played responses to Q37 for two users sharing the same ISTP label.}
\label{tab:same-type-example}
\small
\setlength{\tabcolsep}{5pt}
\renewcommand{\arraystretch}{0.95}
\begin{tabular}{clp{0.55\textwidth}}
\toprule
\textbf{User ID} & \textbf{Role-Played Responses to Q37} & \textbf{Representative Post Evidence} \\
\midrule
3782 & $-3,-3,-2,-3,-3$ & ``I have zero desire to write anything.'' \\
\addlinespace
6170 & $+3,+3,+3,+3,+3$ & ``I'm enveloped in making and watching films. I'm always writing...'' \\
\bottomrule
\end{tabular}
\end{table}

This example complements the aggregate distance analysis by showing that users sharing the same MBTI type can receive substantially different role-played responses when their post evidence differs.

\endgroup

\Needspace{16\baselineskip}
\begingroup
\ifshowchanges\color{yifancolor}\fi

\section{Additional Experiments}
\label{app:additional-experiments}

\subsection{Comparison with EERPD on \textsc{Pandora}}
\label{app:eerpd-pandora}

To complement the inference-time LLM comparisons in Section~3.3, we further evaluate EERPD on the \textsc{Pandora} dataset using the same user-level data split as ROME. Table~\ref{tab:eerpd-pandora} reports the Macro-F1 scores for all four MBTI dimensions.

\begin{table}[H]
\centering
\ifshowchanges\color{yifancolor}\fi
\caption{Comparison with EERPD on the \textsc{Pandora} dataset.}
\label{tab:eerpd-pandora}
\small
\setlength{\tabcolsep}{4pt}
\renewcommand{\arraystretch}{0.9}
\begin{tabular}{lccccc}
\toprule
\textbf{Methods} & \textbf{I/E} & \textbf{S/N} & \textbf{T/F} & \textbf{P/J} & \textbf{Avg} \\
\midrule
EERPD & 62.14 & 56.98 & 70.68 & 60.57 & 62.59 \\
\textbf{ROME} & \textbf{74.93} & \textbf{95.45} & \textbf{82.21} & \textbf{78.22} & \textbf{82.70} \\
\bottomrule
\end{tabular}
\end{table}

ROME achieves an average Macro-F1 of 82.70, outperforming EERPD (62.59) by 20.11 points, with higher scores across all four dimensions. One possible explanation lies in the different modeling strategies: EERPD relies on emotion and emotion-regulation cues, which may be less informative for topic-driven or context-dependent Reddit comments. In contrast, ROME learns questionnaire-grounded evidence covering all four personality constructs without relying exclusively on affective information. These results further demonstrate the effectiveness of ROME on the heterogeneous user-generated content in \textsc{Pandora}.

\Needspace{18\baselineskip}
\subsection{Generalization to Big Five Personality Traits}
\label{app:big-five}

To examine the applicability of ROME beyond MBTI, we further evaluate it on the Essays dataset~\cite{pennebaker1999linguistic} under the Big Five personality framework and compare with the reported results of EERPD~\cite{li2025eerpd}. Specifically, we replace the MBTI-style questionnaire with the 44-item Big Five Inventory (BFI-44)~\cite{john1991bfi}, assigning each item to one of five traits: Agreeableness (AGR), Conscientiousness (CON), Extraversion (EXT), Neuroticism (NEU), and Openness (OPN).

\begin{table}[H]
\centering
\ifshowchanges\color{yifancolor}\fi
\caption{Performance comparison on the Essays dataset under the Big Five personality framework.}
\label{tab:big-five}
\small
\setlength{\tabcolsep}{4pt}
\renewcommand{\arraystretch}{0.9}
\begin{tabular}{lcccccc}
\toprule
\textbf{Methods} & \textbf{AGR} & \textbf{CON} & \textbf{EXT} & \textbf{NEU} & \textbf{OPN} & \textbf{Avg} \\
\midrule
EERPD & 65.01 & \textbf{68.64} & 63.02 & 56.00 & 60.93 & 62.72 \\
\textbf{ROME} & \textbf{65.19} & 63.31 & \textbf{65.06} & \textbf{61.69} & \textbf{65.47} & \textbf{64.14} \\
\bottomrule
\end{tabular}
\end{table}

As shown in Table~\ref{tab:big-five}, ROME achieves an average Macro-F1 of 64.14, outperforming EERPD (62.72) by 1.42 points. ROME performs better on four out of five traits, with particularly notable improvements on NEU (+5.69) and OPN (+4.54), although EERPD performs better on CON. These results suggest that the questionnaire-grounded intermediate supervision and Ask--Answer--Detect framework can be extended beyond MBTI to a different personality construct system without changing the main architecture.

\endgroup
\endgroup
\normalcolor

\section{Limitations and Broader Impacts}
\label{app:limitations-impact}
\paragraph{Limitations.}
The current evaluation focuses on personality detection \yifan{using English-language datasets}, and its generalization to other languages, \yifan{downstream tasks, }or clinical scenarios remains an important direction for future work. 
Nevertheless, this limitation is orthogonal to our core methodological contributions.

\paragraph{Broader impacts.}
ROME has the potential to benefit personality-aware web applications by substantially improving the interpretability and transparency of personality detection. By producing questionnaire-level evidence, it enables researchers and practitioners to trace predictions back to specific psychometric items, offering a more structured and theoretically grounded view of user traits. Beyond application performance, ROME contributes to more accessible and efficient AI systems. Despite its strong performance, ROME relies on a relatively lightweight backbone model, making it computationally efficient and easier to deploy in real-world settings with limited resources. This efficiency may facilitate broader adoption in academic and industry applications, especially in scenarios where large-scale LLM deployment is impractical or cost-prohibitive.

At the same time, personality detection from user data raises concerns regarding privacy, profiling, and potential misuse. We encourage responsible deployment with appropriate safeguards, including transparency to users, informed consent where applicable, and restrictions on sensitive or high-stakes use cases.

\newpage
\input{checklist.tex}
\end{document}

%% file: checklist.tex
\section*{NeurIPS Paper Checklist}
\begin{enumerate}

\item {\bf Claims}
    \item[] Question: Do the main claims made in the abstract and introduction accurately reflect the paper's contributions and scope?
    \item[] Answer: \answerYes{}
    \item[] Justification: The main claims in the abstract and introduction are supported by the proposed Ask--Answer--Detect framework, the question-conditioned MoE architecture, and the experimental results reported in Section~\ref{subsec:overall-results}. The claimed improvements are explicitly reported on the \textsc{Kaggle} and \textsc{Pandora} datasets.
    \item[] Guidelines:
    \begin{itemize}
        \item The answer \answerNA{} means that the abstract and introduction do not include the claims made in the paper.
        \item The abstract and/or introduction should clearly state the claims made, including the contributions made in the paper and important assumptions and limitations. A \answerNo{} or \answerNA{} answer to this question will not be perceived well by the reviewers. 
        \item The claims made should match theoretical and experimental results, and reflect how much the results can be expected to generalize to other settings. 
        \item It is fine to include aspirational goals as motivation as long as it is clear that these goals are not attained by the paper. 
    \end{itemize}

\item {\bf Limitations}
    \item[] Question: Does the paper discuss the limitations of the work performed by the authors?
    \item[] Answer: \answerYes{}
    \item[] Justification: The limitations of ROME are discussed in Appendix~\ref{app:limitations-impact}.
    \item[] Guidelines:
    \begin{itemize}
        \item The answer \answerNA{} means that the paper has no limitation while the answer \answerNo{} means that the paper has limitations, but those are not discussed in the paper. 
        \item The authors are encouraged to create a separate ``Limitations'' section in their paper.
        \item The paper should point out any strong assumptions and how robust the results are to violations of these assumptions (e.g., independence assumptions, noiseless settings, model well-specification, asymptotic approximations only holding locally). The authors should reflect on how these assumptions might be violated in practice and what the implications would be.
        \item The authors should reflect on the scope of the claims made, e.g., if the approach was only tested on a few datasets or with a few runs. In general, empirical results often depend on implicit assumptions, which should be articulated.
        \item The authors should reflect on the factors that influence the performance of the approach. For example, a facial recognition algorithm may perform poorly when image resolution is low or images are taken in low lighting. Or a speech-to-text system might not be used reliably to provide closed captions for online lectures because it fails to handle technical jargon.
        \item The authors should discuss the computational efficiency of the proposed algorithms and how they scale with dataset size.
        \item If applicable, the authors should discuss possible limitations of their approach to address problems of privacy and fairness.
        \item While the authors might fear that complete honesty about limitations might be used by reviewers as grounds for rejection, a worse outcome might be that reviewers discover limitations that aren't acknowledged in the paper. The authors should use their best judgment and recognize that individual actions in favor of transparency play an important role in developing norms that preserve the integrity of the community. Reviewers will be specifically instructed to not penalize honesty concerning limitations.
    \end{itemize}

\item {\bf Theory assumptions and proofs}
    \item[] Question: For each theoretical result, does the paper provide the full set of assumptions and a complete (and correct) proof?
    \item[] Answer: \answerNA{}
    \item[] Justification: The paper does not present formal theoretical results, theorems, or proofs. The proposed framework is evaluated empirically through experiments. 
    \item[] Guidelines:
    \begin{itemize}
        \item The answer \answerNA{} means that the paper does not include theoretical results. 
        \item All the theorems, formulas, and proofs in the paper should be numbered and cross-referenced.
        \item All assumptions should be clearly stated or referenced in the statement of any theorems.
        \item The proofs can either appear in the main paper or the supplemental material, but if they appear in the supplemental material, the authors are encouraged to provide a short proof sketch to provide intuition. 
        \item Inversely, any informal proof provided in the core of the paper should be complemented by formal proofs provided in appendix or supplemental material.
        \item Theorems and Lemmas that the proof relies upon should be properly referenced. 
    \end{itemize}

    \item {\bf Experimental result reproducibility}
    \item[] Question: Does the paper fully disclose all the information needed to reproduce the main experimental results of the paper to the extent that it affects the main claims and/or conclusions of the paper (regardless of whether the code and data are provided or not)?
    \item[] Answer: \answerYes{}
    \item[] Justification: The paper describes the model architecture in Section~\ref{sec:method}, the experimental protocol in Section~\ref{subsec:overall-results}, and implementation details in Appendix~\ref{app:imple}. The code repository is also provided in the abstract to support reproducibility.
    \item[] Guidelines:
    \begin{itemize}
        \item The answer \answerNA{} means that the paper does not include experiments.
        \item If the paper includes experiments, a \answerNo{} answer to this question will not be perceived well by the reviewers: Making the paper reproducible is important, regardless of whether the code and data are provided or not.
        \item If the contribution is a dataset and\slash or model, the authors should describe the steps taken to make their results reproducible or verifiable. 
        \item Depending on the contribution, reproducibility can be accomplished in various ways. For example, if the contribution is a novel architecture, describing the architecture fully might suffice, or if the contribution is a specific model and empirical evaluation, it may be necessary to either make it possible for others to replicate the model with the same dataset, or provide access to the model. In general. releasing code and data is often one good way to accomplish this, but reproducibility can also be provided via detailed instructions for how to replicate the results, access to a hosted model (e.g., in the case of a large language model), releasing of a model checkpoint, or other means that are appropriate to the research performed.
        \item While NeurIPS does not require releasing code, the conference does require all submissions to provide some reasonable avenue for reproducibility, which may depend on the nature of the contribution. For example
        \begin{enumerate}
            \item If the contribution is primarily a new algorithm, the paper should make it clear how to reproduce that algorithm.
            \item If the contribution is primarily a new model architecture, the paper should describe the architecture clearly and fully.
            \item If the contribution is a new model (e.g., a large language model), then there should either be a way to access this model for reproducing the results or a way to reproduce the model (e.g., with an open-source dataset or instructions for how to construct the dataset).
            \item We recognize that reproducibility may be tricky in some cases, in which case authors are welcome to describe the particular way they provide for reproducibility. In the case of closed-source models, it may be that access to the model is limited in some way (e.g., to registered users), but it should be possible for other researchers to have some path to reproducing or verifying the results.
        \end{enumerate}
    \end{itemize}

\item {\bf Open access to data and code}
    \item[] Question: Does the paper provide open access to the data and code, with sufficient instructions to faithfully reproduce the main experimental results, as described in supplemental material?
    \item[] Answer: \answerYes{}
    \item[] Justification: The paper provides a public GitHub repository (https://github.com/JasmineLyf/ROME), linked in the abstract, to support reproducibility. The experiments use existing personality detection datasets, with implementation details and data splits described in Appendix~\ref{app:imple}.
    \item[] Guidelines:
    \begin{itemize}
        \item The answer \answerNA{} means that paper does not include experiments requiring code.
        \item Please see the NeurIPS code and data submission guidelines (\url{https://neurips.cc/public/guides/CodeSubmissionPolicy}) for more details.
        \item While we encourage the release of code and data, we understand that this might not be possible, so \answerNo{} is an acceptable answer. Papers cannot be rejected simply for not including code, unless this is central to the contribution (e.g., for a new open-source benchmark).
        \item The instructions should contain the exact command and environment needed to run to reproduce the results. See the NeurIPS code and data submission guidelines (\url{https://neurips.cc/public/guides/CodeSubmissionPolicy}) for more details.
        \item The authors should provide instructions on data access and preparation, including how to access the raw data, preprocessed data, intermediate data, and generated data, etc.
        \item The authors should provide scripts to reproduce all experimental results for the new proposed method and baselines. If only a subset of experiments are reproducible, they should state which ones are omitted from the script and why.
        \item At submission time, to preserve anonymity, the authors should release anonymized versions (if applicable).
        \item Providing as much information as possible in supplemental material (appended to the paper) is recommended, but including URLs to data and code is permitted.
    \end{itemize}

\item {\bf Experimental setting/details}
    \item[] Question: Does the paper specify all the training and test details (e.g., data splits, hyperparameters, how they were chosen, type of optimizer) necessary to understand the results?
    \item[] Answer: \answerYes{}
    \item[] Justification: The paper specifies the evaluated datasets, train/validation/test split, evaluation metric, baselines, and ablation settings in Section~\ref{subsec:overall-results} and Section~\ref{subsec:ablation}. The training procedure, hyperparameters, model size, and computational setup are further provided in Appendix~\ref{app:imple}.
    \item[] Guidelines:
    \begin{itemize}
        \item The answer \answerNA{} means that the paper does not include experiments.
        \item The experimental setting should be presented in the core of the paper to a level of detail that is necessary to appreciate the results and make sense of them.
        \item The full details can be provided either with the code, in appendix, or as supplemental material.
    \end{itemize}

\item {\bf Experiment statistical significance}
    \item[] Question: Does the paper report error bars suitably and correctly defined or other appropriate information about the statistical significance of the experiments?
    \item[] Answer: \answerYes{}
    \item[] Justification: Our results are reported as the mean and standard deviation over five runs with different random seeds.
    \item[] Guidelines:
    \begin{itemize}
        \item The answer \answerNA{} means that the paper does not include experiments.
        \item The authors should answer \answerYes{} if the results are accompanied by error bars, confidence intervals, or statistical significance tests, at least for the experiments that support the main claims of the paper.
        \item The factors of variability that the error bars are capturing should be clearly stated (for example, train/test split, initialization, random drawing of some parameter, or overall run with given experimental conditions).
        \item The method for calculating the error bars should be explained (closed form formula, call to a library function, bootstrap, etc.)
        \item The assumptions made should be given (e.g., Normally distributed errors).
        \item It should be clear whether the error bar is the standard deviation or the standard error of the mean.
        \item It is OK to report 1-sigma error bars, but one should state it. The authors should preferably report a 2-sigma error bar than state that they have a 96\% CI, if the hypothesis of Normality of errors is not verified.
        \item For asymmetric distributions, the authors should be careful not to show in tables or figures symmetric error bars that would yield results that are out of range (e.g., negative error rates).
        \item If error bars are reported in tables or plots, the authors should explain in the text how they were calculated and reference the corresponding figures or tables in the text.
    \end{itemize}

\item {\bf Experiments compute resources}
    \item[] Question: For each experiment, does the paper provide sufficient information on the computer resources (type of compute workers, memory, time of execution) needed to reproduce the experiments?
    \item[] Answer: \answerYes{}
    \item[] Justification: Appendix~\ref{app:imple} reports the computational setup, including the use of a single NVIDIA RTX 5880 Ada GPU, the use of the ChatGPT API during preprocessing, and the approximate runtime of the full training pipeline.
    \item[] Guidelines:
    \begin{itemize}
        \item The answer \answerNA{} means that the paper does not include experiments.
        \item The paper should indicate the type of compute workers CPU or GPU, internal cluster, or cloud provider, including relevant memory and storage.
        \item The paper should provide the amount of compute required for each of the individual experimental runs as well as estimate the total compute. 
        \item The paper should disclose whether the full research project required more compute than the experiments reported in the paper (e.g., preliminary or failed experiments that didn't make it into the paper). 
    \end{itemize}
    
\item {\bf Code of ethics}
    \item[] Question: Does the research conducted in the paper conform, in every respect, with the NeurIPS Code of Ethics \url{https://neurips.cc/public/EthicsGuidelines}?
    \item[] Answer: \answerYes{}
    \item[] Justification: The research uses existing publicly available personality datasets and does not involve newly recruited human subjects, interventions, or attempts to re-identify individuals. We also discuss potential privacy and profiling risks and responsible deployment considerations in the Broader Impacts section.

    \item[] Guidelines:
    \begin{itemize}
        \item The answer \answerNA{} means that the authors have not reviewed the NeurIPS Code of Ethics.
        \item If the authors answer \answerNo, they should explain the special circumstances that require a deviation from the Code of Ethics.
        \item The authors should make sure to preserve anonymity (e.g., if there is a special consideration due to laws or regulations in their jurisdiction).
    \end{itemize}

\item {\bf Broader impacts}
    \item[] Question: Does the paper discuss both potential positive societal impacts and negative societal impacts of the work performed?
    \item[] Answer: \answerYes{}
    \item[] Justification: The broader impacts are discussed in Appendix~\ref{app:limitations-impact}.
    \item[] Guidelines:
    \begin{itemize}
        \item The answer \answerNA{} means that there is no societal impact of the work performed.
        \item If the authors answer \answerNA{} or \answerNo, they should explain why their work has no societal impact or why the paper does not address societal impact.
        \item Examples of negative societal impacts include potential malicious or unintended uses (e.g., disinformation, generating fake profiles, surveillance), fairness considerations (e.g., deployment of technologies that could make decisions that unfairly impact specific groups), privacy considerations, and security considerations.
        \item The conference expects that many papers will be foundational research and not tied to particular applications, let alone deployments. However, if there is a direct path to any negative applications, the authors should point it out. For example, it is legitimate to point out that an improvement in the quality of generative models could be used to generate Deepfakes for disinformation. On the other hand, it is not needed to point out that a generic algorithm for optimizing neural networks could enable people to train models that generate Deepfakes faster.
        \item The authors should consider possible harms that could arise when the technology is being used as intended and functioning correctly, harms that could arise when the technology is being used as intended but gives incorrect results, and harms following from (intentional or unintentional) misuse of the technology.
        \item If there are negative societal impacts, the authors could also discuss possible mitigation strategies (e.g., gated release of models, providing defenses in addition to attacks, mechanisms for monitoring misuse, mechanisms to monitor how a system learns from feedback over time, improving the efficiency and accessibility of ML).
    \end{itemize}
    
\item {\bf Safeguards}
    \item[] Question: Does the paper describe safeguards that have been put in place for responsible release of data or models that have a high risk for misuse (e.g., pre-trained language models, image generators, or scraped datasets)?
    \item[] Answer: \answerNA{}
    \item[] Justification: The paper does not release a high-risk pretrained model, generative system, or newly scraped dataset. The released materials are intended to reproduce the reported experiments rather than to enable open-ended generation or deployment.
    \item[] Guidelines:
    \begin{itemize}
        \item The answer \answerNA{} means that the paper poses no such risks.
        \item Released models that have a high risk for misuse or dual-use should be released with necessary safeguards to allow for controlled use of the model, for example by requiring that users adhere to usage guidelines or restrictions to access the model or implementing safety filters. 
        \item Datasets that have been scraped from the Internet could pose safety risks. The authors should describe how they avoided releasing unsafe images.
        \item We recognize that providing effective safeguards is challenging, and many papers do not require this, but we encourage authors to take this into account and make a best faith effort.
    \end{itemize}

\item {\bf Licenses for existing assets}
    \item[] Question: Are the creators or original owners of assets (e.g., code, data, models), used in the paper, properly credited and are the license and terms of use explicitly mentioned and properly respected?
    \item[] Answer: \answerYes{}
    \item[] Justification: The creators or original owners of assets (e.g., code, data, models), used in the paper, properly credited and are the license and terms of use explicitly mentioned and properly respected.
    \item[] Guidelines:
    \begin{itemize}
        \item The answer \answerNA{} means that the paper does not use existing assets.
        \item The authors should cite the original paper that produced the code package or dataset.
        \item The authors should state which version of the asset is used and, if possible, include a URL.
        \item The name of the license (e.g., CC-BY 4.0) should be included for each asset.
        \item For scraped data from a particular source (e.g., website), the copyright and terms of service of that source should be provided.
        \item If assets are released, the license, copyright information, and terms of use in the package should be provided. For popular datasets, \url{paperswithcode.com/datasets} has curated licenses for some datasets. Their licensing guide can help determine the license of a dataset.
        \item For existing datasets that are re-packaged, both the original license and the license of the derived asset (if it has changed) should be provided.
        \item If this information is not available online, the authors are encouraged to reach out to the asset's creators.
    \end{itemize}

\item {\bf New assets}
    \item[] Question: Are new assets introduced in the paper well documented and is the documentation provided alongside the assets?
    \item[] Answer: \answerYes{}
    \item[] Justification: The paper introduces code and generated questionnaire-level evidence for the proposed Ask--Answer--Detect framework. The generation protocol, question-to-construct assignment, training procedure, and implementation details are documented in Section~\ref{sec:method} and Appendix~\ref{app:imple}.
    \item[] Guidelines:
    \begin{itemize}
        \item The answer \answerNA{} means that the paper does not release new assets.
        \item Researchers should communicate the details of the dataset\slash code\slash model as part of their submissions via structured templates. This includes details about training, license, limitations, etc. 
        \item The paper should discuss whether and how consent was obtained from people whose asset is used.
        \item At submission time, remember to anonymize your assets (if applicable). You can either create an anonymized URL or include an anonymized zip file.
    \end{itemize}

\item {\bf Crowdsourcing and research with human subjects}
    \item[] Question: For crowdsourcing experiments and research with human subjects, does the paper include the full text of instructions given to participants and screenshots, if applicable, as well as details about compensation (if any)? 
    \item[] Answer: \answerNA{}
    \item[] Justification: The paper does not involve crowdsourcing experiments or newly recruited human subjects. All experiments are conducted on existing personality detection datasets.
    \item[] Guidelines:
    \begin{itemize}
        \item The answer \answerNA{} means that the paper does not involve crowdsourcing nor research with human subjects.
        \item Including this information in the supplemental material is fine, but if the main contribution of the paper involves human subjects, then as much detail as possible should be included in the main paper. 
        \item According to the NeurIPS Code of Ethics, workers involved in data collection, curation, or other labor should be paid at least the minimum wage in the country of the data collector. 
    \end{itemize}

\item {\bf Institutional review board (IRB) approvals or equivalent for research with human subjects}
    \item[] Question: Does the paper describe potential risks incurred by study participants, whether such risks were disclosed to the subjects, and whether Institutional Review Board (IRB) approvals (or an equivalent approval/review based on the requirements of your country or institution) were obtained?
    \item[] Answer: \answerNA{}
    \item[] Justification: The paper does not involve newly recruited human subjects, interventions, surveys, or crowdsourcing experiments conducted by the authors. The study uses existing publicly available datasets.
    \item[] Guidelines:
    \begin{itemize}
        \item The answer \answerNA{} means that the paper does not involve crowdsourcing nor research with human subjects.
        \item Depending on the country in which research is conducted, IRB approval (or equivalent) may be required for any human subjects research. If you obtained IRB approval, you should clearly state this in the paper. 
        \item We recognize that the procedures for this may vary significantly between institutions and locations, and we expect authors to adhere to the NeurIPS Code of Ethics and the guidelines for their institution. 
        \item For initial submissions, do not include any information that would break anonymity (if applicable), such as the institution conducting the review.
    \end{itemize}

\item {\bf Declaration of LLM usage}
    \item[] Question: Does the paper describe the usage of LLMs if it is an important, original, or non-standard component of the core methods in this research? Note that if the LLM is used only for writing, editing, or formatting purposes and does \emph{not} impact the core methodology, scientific rigor, or originality of the research, declaration is not required.
    %this research? 
    \item[] Answer: \answerYes{}
    \item[] Justification: LLM usage is a core component of the proposed method. Section~\ref{subsec:ask} describes how a role-playing LLM generates question-level answers in the offline Ask stage, and Appendix~\ref{app:imple} specifies the use of \texttt{gpt-4o-2024-08-06} and related implementation details.
    \item[] Guidelines:
    \begin{itemize}
        \item The answer \answerNA{} means that the core method development in this research does not involve LLMs as any important, original, or non-standard components.
        \item Please refer to our LLM policy in the NeurIPS handbook for what should or should not be described.
    \end{itemize}

\end{enumerate}